\documentclass[11pt]{article}

\usepackage{acl}

\usepackage{times}
\usepackage{latexsym}
\usepackage[T1]{fontenc}
\usepackage[utf8]{inputenc}
\usepackage{microtype}
\usepackage{inconsolata}

\usepackage{graphicx}
\usepackage{subcaption}
\usepackage{booktabs}
\usepackage{multirow}
\usepackage{makecell}
\usepackage{xurl}
\usepackage[table]{xcolor}
\usepackage{float}
\usepackage{fvextra}
\usepackage[most]{tcolorbox}
\usepackage{enumitem}
\usepackage{amsmath}
\usepackage{amssymb}
\usepackage{mathtools}
\usepackage{amsthm}
\usepackage[capitalize,noabbrev]{cleveref}
\usepackage{bm}

\graphicspath{{./}{figures/}}

\definecolor{lightcyan}{RGB}{225, 245, 250}
\definecolor{icmlgreen}{RGB}{232, 245, 233}
\definecolor{orange-web}{RGB}{255, 165, 0}
\definecolor{sagegreen}{RGB}{138, 179, 137}
\definecolor{lemonyellow}{RGB}{255, 247, 0}
\definecolor{skyblue}{RGB}{135, 206, 235}
\definecolor{coral}{RGB}{255, 127, 80}
\definecolor{lavender}{RGB}{230, 230, 250}
\definecolor{mintgreen}{RGB}{152, 255, 152}
\definecolor{peach}{RGB}{255, 218, 185}
\definecolor{steelblue}{RGB}{70, 130, 180}
\definecolor{rosegold}{RGB}{183, 110, 121}
\colorlet{boxcolor}{orange-web}

\newtcolorbox{promptbox}[2][]{
  enhanced jigsaw,breakable,colback=gray!6!white,colframe=black!75,
  colbacktitle=black!75,coltitle=white,fonttitle=\bfseries\sffamily,
  title={#2},title after break={#2 \hfill \textit{(Continued)}},
  arc=1.5mm,boxrule=1pt,bottomrule at break=1pt,toprule at break=1pt,
  pad at break=2mm,#1
}

\newtcolorbox{fancybox}[2][]{
  enhanced jigsaw,
  breakable,
  colback=boxcolor!10,
  colframe=boxcolor!60,
  colbacktitle=boxcolor!80,
  coltitle=white,
  fonttitle=\bfseries\footnotesize,
  title={#2},
  halign title=flush left,
  sharp corners,
  boxrule=0.8pt,
  left=4pt,
  right=4pt,
  top=4pt,
  bottom=4pt,
  before skip=0.8em,
  after skip=0.8em,
  width=\linewidth,
  before upper={\sloppy\setlength{\emergencystretch}{2em}},
  #1
}

\theoremstyle{plain}

\theoremstyle{definition}

\theoremstyle{remark}

\title{Reward-Decomposed Reinforcement Learning for Immersive Video Role-Playing}

\author{
\textbf{Miao Wang\textsuperscript{1,*}},
\textbf{Yuling Shi\textsuperscript{2,*}},
\textbf{Yijiang Li\textsuperscript{3}},
\textbf{Yeheng Chen\textsuperscript{2}},
\textbf{Xiaodong Gu\textsuperscript{2}}, \\
\textbf{Bin Li\textsuperscript{4}},
\textbf{Bo Gao\textsuperscript{5}},
\textbf{Jun Wang\textsuperscript{6}},
\textbf{Zengxin Han\textsuperscript{7}},
\textbf{Jingtong Wu\textsuperscript{6}},
\textbf{Yaduan Ruan\textsuperscript{1,$\dagger$}}
\\[3pt]
\textsuperscript{1}Nanjing University \quad
\textsuperscript{2}Shanghai Jiao Tong University \quad
\textsuperscript{3}University of California, San Diego \\
\textsuperscript{4}Shenzhen Institutes of Advanced Technology, Chinese Academy of Sciences \\
\textsuperscript{5}School of Information Engineering, Beijing Institute of Graphic Communication \\
\textsuperscript{6}Ant International, Ant Group \quad
\textsuperscript{7}Independent Researcher
\\[3pt]
{\small
\textsuperscript{*}Equal contribution. \quad
\textsuperscript{$\dagger$}Correspondence: \texttt{ruanyaduan@nju.edu.cn}
}
}

\begin{document}
\maketitle
\begin{abstract}
Text-based role-playing models can imitate character styles, but often fail to capture scene atmosphere and evolving tension, which are crucial for immersive applications such as VR games and interactive narratives. We study video-grounded role-playing dialogue and introduce EBM-RL (Eye--Brain--Mouth Reinforcement Learning), a decoupled GRPO-based framework that separates observation (\texttt{<perception>}), reasoning (\texttt{<think>}), and utterance generation (\texttt{<answer>}). This design mimics the human See-Think-Speak process, enabling the model to ground dialogue in visual perception before reasoning and response generation. To optimize this See-Think-Speak process, EBM-RL integrates complementary rewards for scene--text alignment, perceptual--cognitive utility, answer faithfulness, and format consistency. Extensive experiments show that EBM-RL substantially outperforms text-only role-playing baselines and comparable-scale VLM baselines, while remaining competitive with stronger larger-scale VLMs on visual grounding and situational persona metrics.
Moreover, EBM-RL demonstrates strong zero-shot transfer to out-of-domain VideoQA benchmarks without additional fine-tuning. We also release an open-source dataset for video-grounded role-playing dialogue. Our code is available at \url{https://github.com/ebm-rl/RP-See-Think-Speak-RL}.
\end{abstract}

\section{Introduction}
The rapid advancement of large language models (LLMs) \cite{zhao2025surveylargelanguagemodels, chang2023surveyevaluationlargelanguage} has accelerated the development of ``anthropomorphic'' agents \cite{park2023generativeagentsinteractivesimulacra, Wang_2024, gao2025s3socialnetworksimulationlarge}, among which role-playing language agents (RPAs) have attracted increasing attention \cite{chen2025oscarsaitheatersurvey}. By collecting dialogue corpora and character profiles from classic films and television works, existing datasets enable models to generate responses that match a target character's personality and tone \cite{Li2023ChatHaruhiRA, Wang2025CoSER}.

\begin{figure}[t]
  \centering
  \includegraphics[width=\columnwidth]{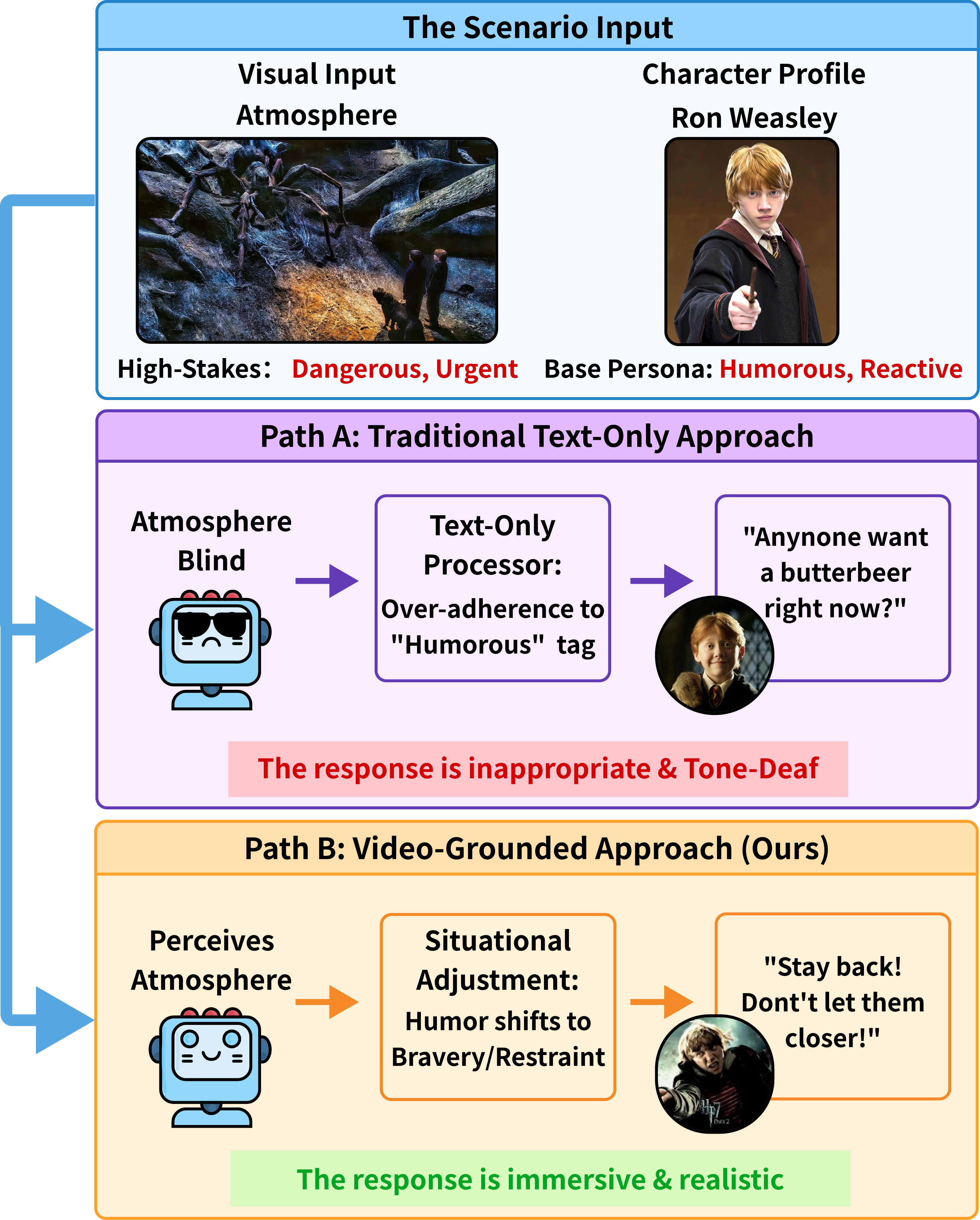}
  \caption{\textbf{From Static Persona to Situational Consistency.} Currently text-only role-playing models (Path A) are ``atmosphere-blind,'' rigidly adhering to humor despite the crisis. In contrast, our immersive video-grounded approach (Path B) perceives visual tension (e.g., giant spiders) and dynamically shifts behavior from humor to restraint, ensuring the character acts appropriately in high-stakes environments.}
  \label{humor}
\end{figure}

Existing role-playing research mainly improves dataset quality, character profiles \cite{Tu2024CharacterEvalAC, Dai2025MMRoleAC, Li2023ChatHaruhiRA}, persona consistency, and generation quality through self-alignment, role-playing-tailored CoT, or reinforcement learning \cite{Lu2024LargeLM, Ji2025EnhancingPC, Liu2025CogDualED}. However, most prior work remains limited to text-only data or static-image-plus-text settings, and open-source video-scenario-oriented role-playing datasets and models remain scarce. As a result, current role-playing agents are largely ``blind'': they rely primarily on dialogue history to judge the situation and often rigidly preserve a character's default speaking style, lacking the visual channel needed to perceive environment, atmosphere, and interaction dynamics.

Although vision-language models (VLMs) have achieved strong progress in visual understanding, localization, and question answering \cite{Maaz2023VideoChatGPTTD, Wang2025InternVL35AO, Bai2025Qwen3VLTR}, they often behave as objective narrators rather than character-specific speakers. They lack stable, distinctive expressions of persona, tone, and values. Moreover, existing multimodal models usually couple ``seeing'' and ``thinking'' in a single reasoning process, making it difficult to determine whether an error arises from visual misperception or flawed reasoning. This coupling may also amplify hallucinations caused by spurious reasoning, reducing the reliability of generated responses.

To address these limitations, we introduce visual perception into role-playing agents so that responses can be conditioned on the current scene, affect, and atmosphere, thereby producing more immersive and context-grounded in-character dialogue. Unlike text-only evaluations that often reward strict adherence to a predefined persona, human-like role-play should support situation-appropriate reactions: for example, a humorous character need not joke in every situation, and showing courage in a crisis may be more faithful than preserving humor, as illustrated in \cref{humor}. This shift from static persona consistency to situational consistency is important for high-fidelity NPC interaction in VR, immersive storytelling, experiential games, and broader open-world interactive agents.

\begin{figure*}[t]
    \centering
    \includegraphics[width=\textwidth]{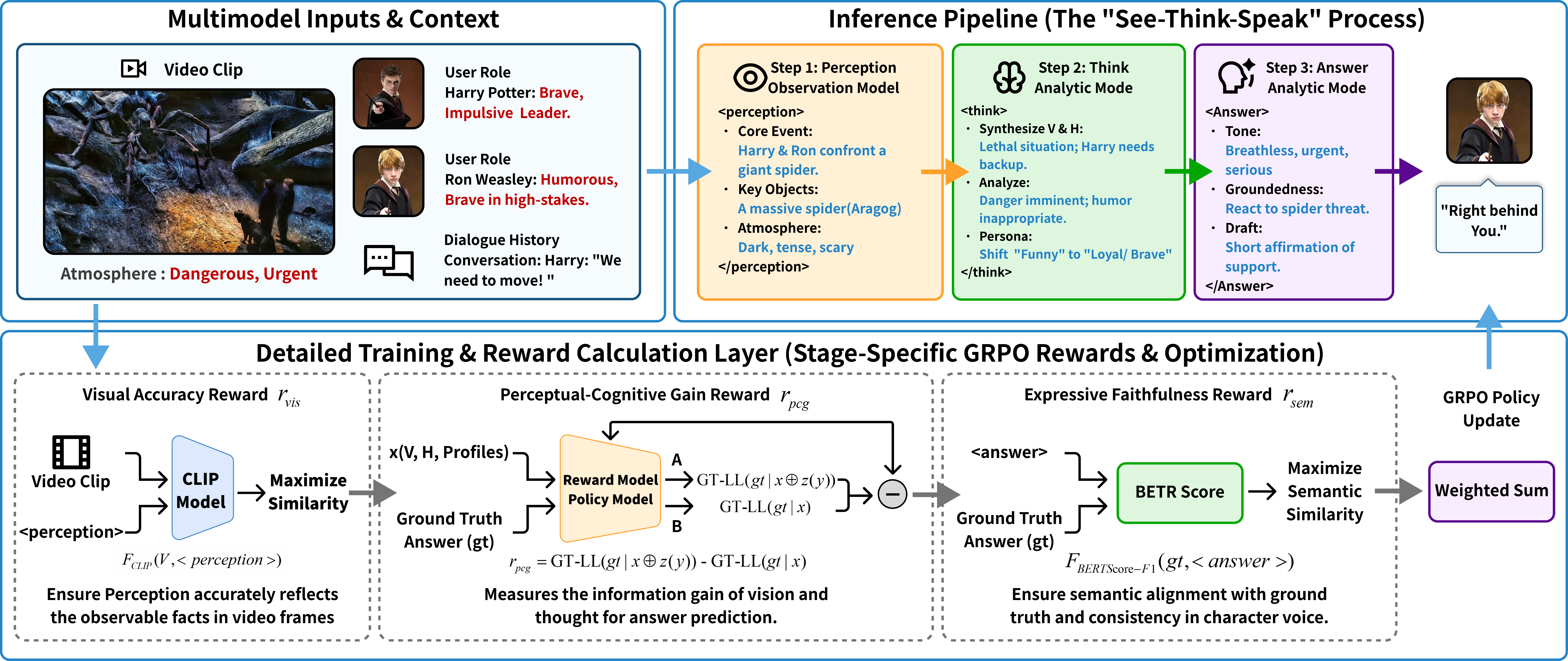} 
    \caption{The Detailed Training Pipeline of the EBM-RL Framework with Stage-Specific GRPO Rewards. This diagram illustrates how each stage of the decoupled ``See-Think-Speak" process is optimized via specific reward mechanisms to ensure visual accuracy, cognitive utility, and expressive faithfulness. $F_{\text{CLIP}}$ and $F_{\text{BERTScore-F1}}$ denote the series of functions calculating the CLIP-based Scene-Text Alignment Reward (\cref{clip reward}) and the open-ended semantic reward (Appendix~\ref{bertscore detail}), respectively.}
    \label{pipeline} 
\end{figure*}

We further decouple seeing and thinking to better approximate human cognition. The agent first observes the scene with its ``eyes,'' then analyzes the perceived visual evidence with its ``mind,'' and finally conditions the utterance on both the situation and the character's persona. Based on this idea, we propose EBM-RL, an Eye-Brain-Mouth reinforcement learning framework that organizes generation into a See-Think-Speak pipeline. To optimize each stage, we design stage-specific GRPO \cite{Shao2024DeepSeekMathPT,Suma2025DeepSeekR1IR} rewards: a CLIP-based \cite{Radford2021LearningTV} scene-text alignment reward for visual grounding, a Perceptual--Cognitive Gain reward for improving perception and reasoning utility, a BERTScore-based \cite{Zhang2019BERTScoreET} semantic reward for faithful and character-consistent utterances, and a format reward for enforcing structured outputs. The pipeline of EBM-RL is visualized in \cref{pipeline}.

Overall, our contributions are threefold: (i) we introduce the first open-source video-grounded role-playing dataset and model for immersive scene-based dialogue; (ii) we propose EBM-RL, a decoupled Eye-Brain-Mouth framework that separates and optimizes seeing, thinking, and speaking with stage-specific GRPO rewards; and (iii) we provide extensive empirical validation against strong VLM and role-playing baselines, together with zero-shot transfer results on out-of-domain VideoQA benchmarks.

\section{Related Works}

\subsection{Role-Playing Language Agents}
Role-playing language agents (RPAs) have attracted increasing attention for emotionally engaging and personalized interactions~\citep{chen2025oscarsaitheatersurvey,Xu2025RPASurvey}. Existing work mainly focuses on dataset and benchmark construction, from the bilingual Harry Potter Dialogue corpus~\citep{Chen2023HPD} to large-scale benchmarks such as LiSCU~\citep{brahman2021letcharacterstellstory}, ToM-in-AMC~\citep{yu2024fewshotcharacterunderstandingmovies}, Character-LLM~\citep{shao2023characterllmtrainableagentroleplaying}, RoleBench~\citep{Wang2024RoleLLM}, CoSER~\citep{Wang2025CoSER}, and OpenCharacter~\citep{Chen2025OpenCharacter}, with evaluation evolving from surface-level metrics to psychological assessments of personality fidelity~\citep{Wang2024InCharacter,Tu2024CharacterEvalAC}. Beyond data, training strategies such as contrastive learning~\citep{Ji2025EnhancingPC} and cognitive-inspired reasoning with reinforcement learning~\citep{Liu2025CogDualED} have been explored to improve persona consistency. However, most prior work remains text-only. Although Video2Roleplay~\citep{Zhang2025Video2Roleplay} introduces video modality into RPAs, it mainly targets first-person lifestyle vlogs and relies on fully automatic synthetic dialogue generation. In contrast, our work studies video-grounded role-playing in cinematic narrative contexts and constructs data from original movie dialogues, preserving authentic colloquialisms and dramatic nuances often lost in synthetic data.

\subsection{Video Large Language Models}
Recent vision-language models have enabled conversational understanding of video content~\citep{Tang2024VideoLLMSurvey}. Early Video LLMs, including VideoChat~\citep{Li2023VideoChat}, Video-ChatGPT~\citep{Maaz2023VideoChatGPTTD}, and Video-LLaMA~\citep{Zhang2023VideoLLaMA}, established foundational architectures~\citep{Zohar2025Apollo} by connecting visual encoders with LLMs through learnable interfaces. Later studies improved spatio-temporal modeling with stronger connectors~\citep{Cheng2024VideoLLaMA2,Zhang2025VideoLLaMA3,chen2025progressive}, enhanced long-video comprehension via dynamic resolution~\citep{Wang2024Qwen2VL,Bai2025Qwen25VL} or context transfer~\citep{Zhang2024LongVA}, and explored fine-grained temporal grounding~\citep{Munasinghe2025VideoGLaMM,Wang2025GroundedVideoLLM}. While effective for objective video understanding and question answering, these models typically behave as neutral narrators and lack character-specific tone and personality. Our work bridges this gap by combining video understanding with role-playing capabilities.

\subsection{Reinforcement Learning for LLMs}
Reinforcement learning has become an effective paradigm for improving LLMs beyond supervised fine-tuning~\citep{Ouyang2022InstructGPT,Rafailov2023DPO}. Group Relative Policy Optimization (GRPO)~\citep{Shao2024DeepSeekMathPT} improves efficiency by removing the critic model and estimating baselines from group scores, while DeepSeek-R1~\citep{Suma2025DeepSeekR1IR} shows that reasoning abilities can emerge through pure RL without human-annotated demonstrations. Subsequent work further improves GRPO scalability~\citep{Yu2025DAPO} and exploration efficiency~\citep{liu2025attention}. In the video domain, GRPO-based reinforcement fine-tuning has been applied to video MLLMs~\citep{Zuo2025VideoChatR1,Feng2025VideoR1,Zhang2025TinyLLaVAVideoR1,Huang2025VideoRFT,cheng2025videoasanswerpredictgeneratevideo}, often using IoU-based or temporal contrastive rewards for spatio-temporal perception and reasoning. However, existing rewards mainly target verifiable tasks with deterministic answers, leaving reward modeling for open-ended generation and vision-text alignment underexplored. Our work addresses this gap with CLIP-based scene-text alignment and perceptual--cognitive gain rewards for immersive video-grounded role-playing.

\section{Task and Dataset}

\subsection{Task Definition}
\paragraph{\textbf{Inputs.}} Given a video clip $V$, user and assistant profiles $(P_u,P_a)$, and dialogue history $H$, the model role-plays as the assistant character and predicts the next utterance $y$. The goal is not only static persona mimicry but also \emph{situational consistency}: the utterance should preserve the assistant's persona while adapting to the visual atmosphere, emotional pressure, and event state in $V$.

\paragraph{\textbf{Output.}} To accommodate our three-stage GRPO training paradigm, the model output must
strictly follow the structured format below:
{\par\centering\small\ttfamily
\verb|<perception>...</perception><think>..|\\
$\hookrightarrow$\verb|..</think><answer>...</answer>|
\par}
The \texttt{<perception>} block reports observable visual facts, including actions, expressions, key objects, and atmosphere. The \texttt{<think>} block integrates these cues with $H$, $P_u$, and $P_a$ to infer the dialogue state and response direction. The \texttt{<answer>} block produces a concise in-character utterance constrained by both persona and scene.

\subsection{Dataset Construction}

Our dataset comprises 37 internationally renowned films from 13 franchises or standalone film series, as summarized in Appendix~\ref{app:film_coverage} and Table~\ref{tab:film_coverage}. To balance authentic colloquial dialogue with interaction diversity, we construct it through two complementary pipelines: a script-grounded pipeline based on original movie dialogues, and an LLM-augmented pipeline that expands interactions under the same visual events. Character profiles are compiled from public sources and structured for our role-playing setting, with examples provided in Appendix~\ref{char profiles}.

\subsubsection{Script-grounded Dialogue Extraction}
\label{raw lines}
We first extract script-grounded samples from original movie dialogues while preserving their temporal alignment with the source films. We use \textit{simple-subtitling}~\citep{huh2025simpleSubtitling} to obtain utterance text, timestamps, and speaker identities, followed by strict manual verification to ensure accurate raw utterances and line-level speaker labels. Based on verified dialogue lines, we construct alternating dialogue sessions $\mathrm{Diag}_{\mathrm{raw}}$ between user and assistant roles under temporal continuity constraints. These sessions are then split into turn-level training samples using the sliding-window strategy of MMRole~\cite{Dai2025MMRoleAC}. Detailed temporal constraints, turn-level splitting, and script-grounded video segmentation procedures are provided in Appendices~\ref{temporal_constraints} and~\ref{video clip method}. We release the construction scripts, users can reproduce the script-grounded pipeline from speaker-attributed \texttt{.srt} files.

\subsubsection{LLM-Augmented Dialogue Expansion}
\label{llm api}
Although script-grounded samples preserve authentic movie-style colloquialisms, they are limited in scale and speaker-pair diversity. We therefore introduce an LLM-augmented pipeline using Gemini~3~Pro. For each video clip, we condition the LLM on a structured visual description, the corresponding raw dialogue $\mathrm{Diag}_{\mathrm{raw}}$, and character profiles to synthesize new dialogues grounded in the same visual event. To avoid simple rewriting of the canonical script, we keep one original interlocutor as the assistant role and re-sample the user role, including a special \textit{user fan} role. We further adopt an \textit{Off-Screen Witness Assumption}, where dialogue participants may be off-camera but are physically present and perceive the event, enabling open-ended and cross-franchise role-playing. Details of the augmentation pipeline, user-role resampling, special roles, and task assumption are provided in Appendix~\ref{app:data_aug}.


\subsubsection{Leakage Prevention and Dataset Split}
To prevent target-response leakage in script-grounded samples, we segment each input video clip to cover only the dialogue history and exclude frames overlapping with the target response. For LLM-generated dialogues, whose target utterances are newly synthesized rather than directly tied to original timestamps, timestamp-based truncation alone is insufficient. We therefore crop out the subtitle region from all video clips to reduce leakage from subtitles or timestamp errors. Across both pipelines, all samples derived from the same video clip are assigned a unified \emph{session id}, and train/test splits are performed by \emph{session id} to avoid contextual leakage. In total, we construct approximately 34k samples. Detailed dataset statistics are reported in Appendix~\ref{dataset} and Table~\ref{table: dataset}, with video duration distributions provided in the same appendix.

\section{Method}
\label{reward method}

\subsection{Overview and Notation}
\label{method overview}

Given an input tuple $x=(V,P_u,P_a,H)$ consisting of a video clip $V$, user/assistant profiles $(P_u,P_a)$, and a dialogue history $H$, the policy $\pi_\theta$ generates a structured completion
\begin{equation}
\begin{aligned}
    y ={}& \texttt{<perception>} y^{v} \texttt{</perception>}
           \texttt{<think>} \\
         & \hookrightarrow y^{t}\texttt{</think>}
           \texttt{<answer>} y^{a} \texttt{</answer>}.
\end{aligned}
\end{equation}
Here, $(y^{v},y^{t},y^{a})$ denote perception, analysis, and answer segments, extracted by $\mathrm{Ext}_{v}$, $\mathrm{Ext}_{t}$, and $\mathrm{Ext}_{a}$; missing or empty segments are treated as $\emptyset$ and receive zero reward.

For each input $x$, GRPO samples a group of $G$ completions $\{y_g\}_{g=1}^{G} \sim \pi_\theta(\cdot \mid x)$. Let $y^{*}$ denote the reference structured output for $x$, and $a^{*}=\mathrm{Ext}_{a}(y^{*})$ be its ground-truth answer segment. Each completion $y_g$ is evaluated by a 4D reward vector
\begin{equation}
\begin{aligned}
\mathbf{r}_g
=
\big[&r_{\text{fmt}}(y_g),\;
      r_{\text{sem}}(y_g^{a},a^{*}),\\
     &r_{\text{vis}}(V,y_g^{v}),\;
      r_{\text{pcg}}(x,y_g^{v},y_g^{t})
\big].
\end{aligned}
\end{equation}
Any reward requiring a segment is set to $0$ if the required segment equals $\emptyset$, preventing malformed outputs from being rewarded.

\subsection{CLIP-based Scene--Text Alignment Reward}
\label{clip reward}

To encourage evidence-grounded perception, we define a CLIP-based reward between the generated perception text $y^{v}$ and video $V$. We uniformly sample $N$ representative frames from $V$ and precompute normalized CLIP image embeddings:
\begin{equation}
v_j = \frac{f_{\text{img}}(f_j)}{\|f_{\text{img}}(f_j)\|_2} \in \mathbb{R}^d,\quad j=1,\dots,N.
\end{equation}
For $y^{v}$, we compute the normalized CLIP text embedding:
\begin{equation}
u = \frac{f_{\text{text}}(y^{v})}{\|f_{\text{text}}(y^{v})\|_2} \in \mathbb{R}^d,
\end{equation}
so the frame--text similarity is $s_j = v_j^\top u$, i.e., cosine similarity after normalization.

We use two aggregation variants. \textbf{(i) CLIP-Max} takes the maximum frame similarity:
\begin{equation}
r_{\text{vis}}^{\max}(V,y^{v}) = \max_{j\in\{1,\dots,N\}} v_j^\top u.
\end{equation}
\textbf{(ii) CLIP-SentTopK} splits $y^{v}$ into $M$ sentences $\{s_m\}_{m=1}^{M}$, embeds each sentence as $u_m$, and forms $S \in \mathbb{R}^{N\times M}$ with $S_{j,m}=v_j^\top u_m$. For each sentence, we average the top-$K$ frame similarities and then average over sentences:
\begin{equation}
\begin{aligned}
    r(s_m)
    &= \frac{1}{K}
       \sum_{j \in \mathrm{TopK}(S_{:,m})} S_{j,m}, \\
    r_{\text{vis}}^{\text{TopK}}(V,y^{v})
    &= \frac{1}{M}\sum_{m=1}^{M} r(s_m).
\end{aligned}
\end{equation}
where $K=\max(1,\lfloor \alpha N \rfloor)$ and $\alpha\in(0,1]$ is a hyperparameter. Frame sampling, sentence splitting, and $\alpha$ are detailed in Appendix~\ref{clip topk detail}.

\subsection{Perceptual--Cognitive Gain Reward}
\label{pcg reward}


The PCG reward encourages the intermediate perception--cognition blocks $(y^{v},y^{t})$ to increase support for the ground-truth next utterance. Let $d^{gt}=(d_1,\dots,d_T)$ denote the ground-truth answer tokens, and let $c$ denote the conditioning context given to a frozen reference policy $\pi_{\text{ref}}$. We define
\begin{equation}
\mathrm{GT\text{-}LL}(d^{gt}\mid c)=\frac{1}{T}\sum_{t=1}^{T}\log \pi_{\text{ref}}(d_t\mid d_{<t},c).
\end{equation}

To prevent answer-cue tokens in the reasoning block from directly increasing the PCG likelihood-gain term, we compute PCG on a lexically cleaned intermediate block:
\begin{equation}
\bar{z}(y)=y^{v}\oplus \mathrm{Clean}_{\mathrm{lex}}(y^{t},y^{a}),
\end{equation}
where $\mathrm{Clean}_{\mathrm{lex}}$ removes answer-copying reasoning sentences. We also subtract $L_{\mathrm{copy}}(y^{t},y^{a})$, which measures lexical copying between the generated reasoning and answer segments.

The PCG reward is then defined as
\begin{equation}
\resizebox{0.96\columnwidth}{!}{$
\begin{aligned}
r_{\text{pcg}}(x,y^{v},y^{t})
={}&
\mathrm{GT\text{-}LL}\bigl(d^{gt}\mid x\oplus \bar{z}(y)\bigr) \\
&-
\mathrm{GT\text{-}LL}\bigl(d^{gt}\mid x\bigr)
-
L_{\mathrm{copy}}(y^{t},y^{a}).
\end{aligned}
$}
\end{equation}
The detailed lexical cleaning and copy-penalty computation are provided in Appendix~\ref{pcg lexical safeguard}.

\subsection{Semantic and Format Rewards}
\label{bertscore sec}
\label{acc reward}

For open-ended role-play responses, we use BERTScore to measure semantic similarity between the predicted and ground-truth answer segments. Let $\hat{a}=y^{a}$ and $a^{*}=\mathrm{Ext}_{a}(y^{*})$. We define
\begin{equation}
r_{\text{sem}}(\hat{a},a^{*})
=
\text{clip}_{[0,1]}\Big( \text{BERTScore-F1}(\hat{a}, a^{*}) \Big).
\end{equation}
The token-level BERTScore definition and encoder details are provided in Appendix~\ref{bertscore detail}.

\subsection{Dense Format Reward}
We adopt a dense format reward $r_{\text{fmt}}(y)$ to enforce the required structure. It combines tag existence constraints for \texttt{<perception>}, \texttt{</perception>}, \texttt{<think>}, \texttt{</think>}, \texttt{<answer>}, and \texttt{</answer>}; structural order consistency; and boundary constraints requiring the output to start with \texttt{<perception>} and end with \texttt{</answer>}. The score can be negative, distinguishing partially correct outputs from invalid ones. The exact scoring function is provided in Appendix~\ref{dense reward}.

\subsection{Reward Aggregation and Optimization Signal}
\label{combine rewards}

Since reward components have different scales, we perform per-dimension Z-score normalization over the sampled group. Let $r_{g,k}$ be the raw score of sample $g$ on reward dimension $k\in\{\text{fmt},\text{sem},\text{vis},\text{pcg}\}$. We compute
\begin{equation}
\mu_k = \frac{1}{G}\sum_{g=1}^{G} r_{g,k},\quad
\sigma_k = \sqrt{\frac{1}{G}\sum_{g=1}^{G} (r_{g,k}-\mu_k)^2},
\end{equation}
and normalize rewards as
\begin{equation}
\tilde{r}_{g,k} = \frac{r_{g,k}-\mu_k}{\sigma_k+\varepsilon},\quad \varepsilon=10^{-6}.
\end{equation}
The scalar advantage is then
\begin{equation}
A_g = \sum_{k} w_k \tilde{r}_{g,k},
\end{equation}
where the weight vector is set to $\mathbf{w}=[w_{\text{fmt}}, \allowbreak w_{\text{sem}}, \allowbreak w_{\text{vis}}, \allowbreak w_{\text{pcg}}] = \allowbreak [1.0, 1.0, 0.8, 0.8]$. The weight choice is analyzed in \cref{Ablation two}. This advantage is used for GRPO training with a KL regularizer; the full objective is given in Appendix~\ref{grpo}.

\section{Experiments}
\subsection{Experimental Setup}

\paragraph{\textbf{Training.}}
We partition the data by session id to prevent clip-level leakage caused by overlapping training/test samples from the same session topic; the detailed split is shown in \cref{table: dataset}. To mitigate format collapse of the structured output during RL, we additionally construct 2.3k high-quality CoT exemplars using Gemini~3~Pro and perform stage-1 SFT for nearly three epochs as a warm start. Details of CoT data construction are provided in Appendix~\ref{sft cot}. We use Qwen2.5-VL-7B-Instruct as the base model. In the reinforcement learning stage, we fine-tune the SFT-initialized model on our constructed training set for one epoch using 8 NVIDIA A100 GPUs.

\paragraph{\textbf{Evaluation and Metrics.}} 

Since video-grounded role-playing requires visual grounding, situational adaptation, and natural dialogue, we adapt existing role-playing evaluation paradigms~\citep{Wang2025CoSER, Liu2025CogDualED} into three LLM-judge metrics aligned with the ``See-Think-Speak'' pipeline: \textbf{Visual Evidence Grounding (VEG)}, which assesses visual perception and hallucination; \textbf{Situational Persona Compatibility (SPC)}, which extends Character Fidelity~\citep{Wang2025CoSER} with environmental risk/stress constraints; and \textbf{Conversational Naturalism (CN)}, which measures human-like colloquial fluency and avoids templated AI-style responses. Since our goal is immersive dialogue rather than storyline continuation, we exclude plot-continuation metrics and use VEG/SPC to verify scene constraints and character consistency. All evaluation prompts are listed in Appendix~\ref{metrics}.

We adopt GPT-5-mini as the LLM judge for automatic evaluation. To reduce potential brand/scale bias, we anonymize model identities during judging (e.g., Model A/B/...) and only remap scores to the actual models after scoring. All reported main results correspond to a single training run.

\paragraph{\textbf{Baselines.}} 

We compare with three categories of baselines. \textbf{(i) General VLMs:} InternVL3--8B/14B/38B--Instruct~\citep{Wang2025InternVL35AO} and Qwen2.5-VL--7B/32B--Instruct~\citep{Bai2025Qwen25VL}, which test visual perception of character emotions, scene atmosphere, and dialogue-driving cues. \textbf{(ii) Text-only role-playing models:} RoleMRC-dpo~\citep{Lu2025RoleMRCAF}, Crab~\citep{He2025CrabAN}, and Haruhi~\citep{haruhizero2024}, which test whether strong persona models can adapt to video-grounded situational constraints. \textbf{(iii) Proprietary frontier models:} GPT-5~\citep{OpenAI2025GPT5} and Gemini-3 Pro~\citep{GoogleDeepMind2025Gemini3}, reported as reference upper-bound systems. We also include Qwen2.5-VL-7B-SFT, fine-tuned on our constructed CoT data,

\begin{table}[!t]
    \centering
  \small
  \setlength{\tabcolsep}{2.0pt}
  \renewcommand{\arraystretch}{1.02}
  \begin{tabular}{@{}lccc|c@{}}
    \toprule
    Model & VEG$\uparrow$ & SPC$\uparrow$ & CN$\uparrow$ & Avg.$\uparrow$ \\
    \midrule
    GPT-5                         & 79.11 & 79.01 & 81.22 & 79.78 \\
    Gemini-3                      & 80.08 & 77.37 & 83.96 & 80.47 \\
    \midrule
    InternVL3-38B-Instruct         & 74.95 & 70.82 & 74.43 & 73.4 \\
    Qwen2.5-VL-32B-Instruct        & 74.61 & 71.01 & 73.32 & 72.98 \\
    InternVL3-14B-Instruct         & 73.87 & 67.75 & 72.78 & 71.47 \\
    \midrule
    InternVL3-8B-Instruct          & 71.85 & 65.53 & 72.76 & 70.05 \\
    Qwen2.5-VL-7B-Instruct         & 70.75 & 65.05 & 71.71 & 69.17 \\
    RoleMRC               & 70.51 & 66.77 & 73.14 & 70.14 \\
    Crab                  & 67.86 & 66.87 & \textbf{75.23} & 69.99 \\
    Haruhi                & 67.09 & 62.09 & 72.38 & 67.19 \\
    Qwen2.5-VL-7B-SFT     & 70.82 & 66.46 & 72.08 & 69.79 \\
    \rowcolor{cyan!18}
    Char-EBM-CLIP-TopK    & \underline{73.47} & \underline{69.94} & 73.69 & \underline{72.37} \\
    \rowcolor{cyan!18}
    Char-EBM-CLIP-Max     & \textbf{74.25} & \textbf{70.37} & \underline{74.78} & \textbf{73.13} \\
    \bottomrule
  \end{tabular}
\caption{Evaluation results on three role-play dimensions. VEG: Visual Evidence Grounding; SPC: Situational Persona Compatibility; CN: Conversational Naturalism. Among comparable-scale models to EBM, the best and second-best results are highlighted in \textbf{bold} and \underline{underlined}, respectively.}
  \label{tab:main_results}
  \vskip -0.1in
\end{table}

\subsection{Main Results}

As shown in Table~\ref{tab:main_results}, GPT-5 and Gemini-3 Pro provide proprietary frontier reference points, while our main comparison focuses on comparable-scale open-source and role-playing baselines. The three metrics correspond to the ``See-Think-Speak'' pipeline. On VEG, text-only role-playing baselines underperform because they lack direct visual perception and must infer scene states from dialogue history. Char-EBM-CLIP-Max achieves VEG parity with the much larger Qwen2.5-VL-32B, suggesting that the RL vision reward improves perception of conversational atmosphere and visible emotional cues. Its advantage over CLIP-SentTopK further indicates that immersive dialogue is often driven by sparse, high-magnitude visual triggers, such as sudden emotional shifts, whereas averaging-based aggregation can dilute these critical signals and is more suitable for dense tasks like video captioning.

The largest gain appears on SPC, where Char-EBM-CLIP-Max reaches 70.37, improving over Qwen2.5-VL-7B by +5.32. As a proxy for the ``thinking'' stage, SPC shows that EBM-RL better synthesizes visual cues with persona traits, maintaining character fidelity in personality, tone, and values while dynamically modulating responses to situational stakes. The weaker SPC of CLIP-SentTopK suggests that visual perception bottlenecks can propagate to downstream reasoning. On CN, our script-derived data promotes concise and fluent human-like dialogue, but our model still slightly trails Crab, likely because Crab is trained on a larger role-playing corpus of 41k samples.

To provide a fairer comparison with text-only role-playing agents, we additionally construct visual-augmented variants of RoleMRC, Crab, and Haruhi by conditioning them on video descriptions generated by Gemini-3-Pro-Preview. As detailed in Appendix~\ref{app:visual_augmented_rp}, visual augmentation consistently improves VEG, but does not always translate into higher SPC, suggesting that simply providing visual descriptions is insufficient for robust visual--persona integration. Char-EBM-CLIP-Max remains the strongest overall model among these comparable systems.


\subsection{Held-out-Film Test}
\label{sec:out-domain}

To evaluate generalization to unseen films and characters within the cinematic role-playing domain, we construct a held-out-film subset whose source films and corresponding characters are entirely excluded from training. This experiment is intended to measure held-out film/character generalization within the same cinematic domain rather than arbitrary open-domain transfer.

\begin{table}[!t]
    \centering
    \small
    \setlength{\tabcolsep}{2.0pt}
    \renewcommand{\arraystretch}{1.02}
    \resizebox{\columnwidth}{!}{%
    \begin{tabular}{@{}lccc|c@{}}
        \toprule
        Model & VEG$\uparrow$ & SPC$\uparrow$ & CN$\uparrow$ & Avg.$\uparrow$ \\
        \midrule
        InternVL3-38B         & 78.90 & 69.85 & 80.30 & 76.35 \\
        Qwen2.5-VL-32B        & 77.83 & 69.76 & 76.33 & 74.64 \\
        InternVL3-14B         & 76.88 & 69.01 & 75.39 & 73.76 \\
        \midrule
        InternVL3-8B          & 71.16 & 61.21 & 71.28 & 67.88 \\
        Qwen2.5-VL-7B         & 73.89 & 62.05 & 72.14 & 69.36 \\
        RoleMRC               & 71.50 & 58.16 & 71.77 & 67.14 \\
        Crab                  & 70.82 & 62.92 & \textbf{75.66} & 69.80 \\
        Haruhi                & 67.09 & 55.89 & 63.08 & 62.02 \\
        Qwen2.5-VL-7B-SFT     & 73.91 & 64.66 & 73.19 & 70.59 \\
        \rowcolor{cyan!18}
        Char-EBM-CLIP-Max     & \textbf{75.73} & \textbf{67.11} & \underline{75.34} & \textbf{72.73} \\
        \bottomrule
    \end{tabular}%
    }
    \caption{Evaluation on held-out films and characters excluded from training within the cinematic role-playing domain.}
    \label{tab:out-of-domain}
    \vskip -0.1in
\end{table}

As shown in Table~\ref{tab:out-of-domain}, Char-EBM-CLIP-Max achieves the best average performance among comparable-scale models, reaching an Avg. score of 72.73. Compared with Qwen2.5-VL-7B, Qwen2.5-VL-7B-SFT, and InternVL3-8B, it improves the average score by 3.37, 2.14, and 4.85 points, respectively. The gain is particularly evident on SPC, where Char-EBM-CLIP-Max outperforms Qwen2.5-VL-7B by 5.06 points. This suggests that the learned See-Think-Speak decomposition helps transfer situational persona reasoning to films and characters unseen during training. Although substantially larger VLMs such as InternVL3-38B still benefit from greater model capacity, EBM-RL demonstrates strong held-out-film and held-out-character generalization within the cinematic role-playing domain.

\subsection{Robustness and Validation}
\label{sec:robustness}

To reduce the sensitivity of score-based LLM-as-a-Judge evaluation, we conduct preference-based pairwise comparison and human-assisted consistency validation. In the pairwise protocol, the judge selects only Win/Loss without numeric scores, and each pair is evaluated with swapped response order to reduce position bias. Table~\ref{tab:pairwise_summary} provides a compact summary, with full Win/Loss/Tie rates, Decisive Win Rates, and exact binomial-test $p$-values in Appendix~\ref{app:pairwise_all} and Table~\ref{tab:pairwise_all}.

Although the average-score gaps in Table~\ref{tab:main_results} are moderate, pairwise results show that Char-EBM-CLIP-Max is consistently preferred across samples and dimensions. It obtains stable Net Win gains over Qwen2.5-VL-7B-Instruct on VEG, SPC, and CN (+24.6/+24.2/+26.0), and is also preferred over the larger InternVL3-14B (+12.6/+8.6/+16.4), suggesting that reward-decomposed training improves role-playing quality beyond simply increasing VLM scale.

For text-only role-playing models, EBM-RL obtains strong gains over RoleMRC and Haruhi, showing that text-only persona modeling is insufficient under video-grounded situational constraints. The comparison with Crab is more nuanced: EBM strongly improves VEG and SPC (+24.8/+24.2), while its CN advantage is marginal and not statistically significant (+1.2), consistent with Table~\ref{tab:main_results}. This supports our claim that EBM-RL mainly improves visual grounding and situational persona adaptation while preserving competitive conversational naturalism. 

Human validation in Appendix~\ref{app:human_llm_validation} further corroborates these findings: human-only pairwise comparisons reproduce the same overall trends, while human and GPT-5-mini evaluations show high system-level correlations and 90.5--97.0\% within-one-tier agreement at the sample level. An additional evaluation with the independent Gemini-3-Flash-Preview judge in Appendix~\ref{app:gemini_judge_validation} further shows highly consistent model rankings across judges, despite metric-specific differences in absolute score calibration.

\begin{table}[!t]
\centering
\small
\setlength{\tabcolsep}{2.8pt}
\renewcommand{\arraystretch}{0.95}
\begin{tabular}{@{}lcc@{}}
\toprule
Opponent & \makecell[c]{EBM Net Win on\\VEG/SPC/CN} & \makecell[c]{Avg. Decisive\\Win} \\
\midrule
Intern-VL-14B-Instruct       & +12.6 / +8.6 / +16.4 & 57.8 \\
Qwen-VL-7B-Instruct       & +24.6 / +24.2 / +26.0 & 65.4 \\
Qwen-VL-7B-SFT   & +19.4 / +16.4 / +18.4 & 60.8 \\
RoleMRC       & +24.0 / +13.8 / +26.0 & 63.0 \\
Crab          & +24.8 / +24.2 / +1.2  & 60.2 \\
Haruhi        & +46.6 / +48.2 / +44.2 & 77.5 \\
\bottomrule
\end{tabular}
\caption{Pairwise preference summary. Net Win is Win rate minus Loss rate. Avg. Decisive Win is averaged over VEG, SPC, and CN after excluding ties.}
\label{tab:pairwise_summary}
\vskip -0.1in
\end{table}

\subsection{Cross-Domain Test}
\label{sec:holoassist_cross_domain}

To further examine generalization beyond cinematic scenes, we adapt HoloAssist~\citep{Wang_2023_ICCV}, an egocentric real-world human collaboration dataset, into a quality-controlled 500-sample cross-domain pilot. HoloAssist differs substantially from our movie benchmark in viewpoint, visual content, and interaction structure: it contains first-person physical manipulation and remote instructional dialogue rather than third-person cinematic narratives. As detailed in Appendix~\ref{app:holoassist_cross_domain}, Char-EBM-CLIP-Max improves over its Qwen2.5-VL-7B base model by +4.66/+2.14/+5.43 on VEG/SPC/CN and +4.08 on average, suggesting that the learned See-Think-Speak capability transfers to substantially different egocentric real-world interactions.

\begin{table}[!t]
    \centering
  \small
  \setlength{\tabcolsep}{1.8pt}
  \renewcommand{\arraystretch}{1.02}
  \begin{tabular}{@{}lccc|c@{}}
    \toprule
    Model & VEG$\uparrow$ & SPC$\uparrow$ & CN$\uparrow$ & Avg.$\uparrow$ \\
    \midrule
    \rowcolor{cyan!18}
    \makecell[l]{Char-EBM-CLIP-Max \\ \scriptsize ($w$: 1.0, 1.0, 0.8, 0.8)} & \textbf{74.25} & 70.37 & \textbf{74.78} & \textbf{73.13} \\
    \midrule
    \makecell[l]{Char-EBM-NO-CLIP \\ \scriptsize ($w$: 1.0, 1.0, 0.8, 0.8)} & 73.03 & 69.71 & 74.57 & 72.44 \\
    \addlinespace[0.2ex]
    \makecell[l]{Char-EBM-NO-PCG \\ \scriptsize ($w$: 1.0, 1.0, 0.8, 0.8)} & 70.92 & 66.53 & 72.24 & 69.9 \\
    \midrule
    \makecell[l]{Char-EBM-CLIP-MAX \\ \scriptsize ($w$: 1.0, 1.0, 1.0, 1.0)} & 74.1 & \textbf{70.54} & 74.69 & 73.11 \\
    \addlinespace[0.2ex]
    \makecell[l]{Char-EBM-CLIP-MAX \\ \scriptsize ($w$: 1.0, 1.2, 1.0, 1.0)} & 71.39 & 67.29 & 72.11 & 70.26 \\
    \bottomrule
  \end{tabular}
\caption{Ablation study on reward modules and reward weights. $w$ denotes the weights of $r_{\text{fmt}}$, $r_{\text{sem}}$, $r_{\text{vis}}$, and $r_{\text{pcg}}$, respectively.}
  \label{tab:Ablation one}
  \vskip -0.1in
\end{table}

\subsection{Ablation Study}
\subsubsection{Effectiveness of Reward Modules.}
\label{ablation one}
Table~\ref{tab:Ablation one} evaluates the contributions of the Scene--Text Alignment reward $r_{\text{vis}}$ and the Perceptual--Cognitive Gain reward $r_{\text{pcg}}$. Removing either component degrades VEG, SPC, and CN, validating the effectiveness of the reward-decomposed pipeline. Removing $r_{\text{vis}}$ mainly hurts VEG and slightly lowers SPC, because weaker visual alignment makes the model less reliable in identifying character emotions and scene atmosphere; such perceptual errors then propagate to downstream persona reasoning. CN is less affected because it is primarily supervised by the semantic reward in the final answer stage.

Removing $r_{\text{pcg}}$ causes a larger performance drop than removing $r_{\text{vis}}$, especially on SPC and CN. This suggests that although the model can often perceive coarse scene atmosphere with the help of $r_{\text{vis}}$, the ``thinking'' stage has a higher optimization ceiling: the model still needs to transform visual evidence into a response plan that is consistent with the dialogue state, character relationship, and situational stakes. Without $r_{\text{pcg}}$, the correct environmental context is less effectively transmitted from perception to the final utterance, weakening both situational persona compatibility and conversational naturalness.

\subsubsection{Reward Weight Sensitivity Analysis}
\label{Ablation two}
We further study the reward weight vector $w=[w_{\text{fmt}},w_{\text{sem}},w_{\text{vis}},w_{\text{pcg}}]$, corresponding to format, semantic, scene--text alignment, and PCG rewards. Increasing $w_{\text{vis}}$ and $w_{\text{pcg}}$ from 0.8 to 1.0 slightly improves SPC, indicating that stronger visual-cognitive supervision enhances the model's ability to reason about scene-conditioned character behavior. However, this setting marginally reduces VEG and CN. One possible reason is that the PCG reward, due to its larger optimization space in the thinking stage, can dominate the total training signal and suppress the influence of scene--text alignment and final-answer semantic rewards.

When the semantic reward is further increased from $[1,1,1,1]$ to $[1,1.2,1,1]$, performance drops across all metrics. This indicates that over-emphasizing final-answer similarity can bias optimization toward surface-level semantic matching, weakening the coordinated See--Think--Speak process. Overall, $w=[1.0,1.0,0.8,0.8]$ achieves the best average performance by balancing visual grounding, cognitive utility, response semantics, and format consistency.

\subsection{Cross-task Zero-shot Transfer to VideoQA}
\label{sec:ood_generalization}


To test whether EBM-RL learns transferable video reasoning rather than role-playing-specific patterns, we evaluate it zero-shot on unseen VideoQA benchmarks without fine-tuning: NExT-QA~\citep{Xiao2021NExTQANP} for real-world daily activities, PororoQA~\citep{Kim2017DeepStoryVS} for animated narratives, and ActivityNet-QA~\citep{Yu2019ActivityNetQAAD} for open-web long videos. These benchmarks differ from our movie role-playing data in task format, scene source, and visual style. As shown in Table~\ref{tab:ood_results}, EBM improves over Qwen2.5-VL-7B-CoT by +1.93, +2.60, and +2.50 on the three benchmarks, suggesting that See-Think-Speak transfers to general video understanding under cross-task and cross-domain shifts.

\begin{table}[!t]
    \centering
  \footnotesize
  \setlength{\tabcolsep}{1.8pt}
  \renewcommand{\arraystretch}{1.04}
  \begin{tabular}{@{}llccc@{}}
    \toprule
    Dataset & Subset & \makecell{Qwen-7B\\VL-CoT} & EBM-Max & $\Delta$ \\
    \midrule
    NExT-QA & CH   & 70.59 & \cellcolor{cyan!18}72.63 & \cellcolor{cyan!18}\textbf{+2.04} \\
    NExT-QA & CW   & 73.84 & \cellcolor{cyan!18}75.22 & \cellcolor{cyan!18}\textbf{+1.38} \\
    NExT-QA & TC   & 70.73 & \cellcolor{cyan!18}73.13 & \cellcolor{cyan!18}\textbf{+2.40} \\
    NExT-QA & TN   & 62.04 & \cellcolor{cyan!18}64.55 & \cellcolor{cyan!18}\textbf{+2.51} \\
    NExT-QA & DL   & 90.06 & \cellcolor{cyan!18}91.30 & \cellcolor{cyan!18}\textbf{+1.24} \\
    NExT-QA & DO   & 82.50 & \cellcolor{cyan!18}84.50 & \cellcolor{cyan!18}\textbf{+2.00} \\
    NExT-QA & Avg. & 74.96 & \cellcolor{cyan!18}76.89 & \cellcolor{cyan!18}\textbf{+1.93} \\
    PororoQA       & 2k    & 48.85 & \cellcolor{cyan!18}51.45 & \cellcolor{cyan!18}\textbf{+2.60} \\
    ActivityNet-QA & Y/N   & 77.40 & \cellcolor{cyan!18}79.90 & \cellcolor{cyan!18}\textbf{+2.50} \\
    \bottomrule
  \end{tabular}
\caption{Zero-shot transfer to external VideoQA benchmarks. Avg. denotes the NExT-QA average; 2k and Y/N denote the PororoQA subset and ActivityNet-QA yes/no subset.}
  \label{tab:ood_results}
  \vskip -0.1in
\end{table}

\section{Conclusion}

We propose EBM-RL, a decoupled reinforcement learning framework for video-grounded role-playing that addresses the lack of ``situational consistency'' in existing role-playing agents. Mimicking the human ``See-Think-Speak'' process, EBM-RL explicitly partitions generation into observation (\texttt{<perception>}), reasoning (\texttt{<think>}), and utterance (\texttt{<answer>}), and optimizes this process with CLIP-based scene--text alignment and Perceptual--Cognitive Gain rewards. Experiments show that EBM-RL improves visual grounding, situational persona adaptation, and conversational authenticity over strong comparable-scale baselines. Moreover, its zero-shot gains on unseen VideoQA benchmarks demonstrate that the learned See-Think-Speak policy transfers beyond movie role-playing to broader video understanding scenarios. Overall, EBM-RL provides a practical path toward immersive NPCs and video-grounded interactive agents.


\section*{Limitations}

This work takes an initial step toward video-grounded role-playing by constructing a benchmark primarily from internationally renowned films. Such scenarios provide rich characters, expressive visual contexts, and high-quality dialogues, making them suitable for studying situational consistency. Nevertheless, future work may further extend the data coverage to real-world role-playing scenarios, such as daily interactions, service-oriented conversations, or user-created immersive scenes, to examine the framework under broader interaction role-playing styles.

In addition, our experiments are conducted in offline video-conditioned settings, where the model generates responses based on pre-collected video clips and dialogue histories. This setting allows controlled evaluation of visual grounding and character consistency, but it does not fully capture the dynamics of real-time interactive VR or game environments. Extending EBM-RL to live human-agent interaction, where visual states, user actions, and dialogue evolve continuously, is an important direction for future work.

\section*{Ethics Statement}

This work advances immersive role-playing by enabling agents to perceive scene atmosphere via a decoupled ``See-Think-Speak'' RL framework. This fosters realistic NPC development for VR and interactive narratives while improving safety by reducing multimodal hallucinations. Regarding data ethics, our movie-based dataset is strictly for non-commercial academic research. To respect copyright and prevent data leakage, we masked subtitles and removed audio tracks. We will limit our open-source release to keyframes or processing scripts rather than raw videos. Finally, we advocate for transparent deployment to mitigate potential risks associated with increasingly anthropomorphic AI.

\section*{Acknowledgments}
This work was supported by the Shenzhen Medical Research Fund (No. D2404001), and in part by the National Natural Science Foundation of China (No. 62472277), the Shanghai East Talents Program (2023-177), the Key Research and Development Program of Guangdong Province (No. 2025B1111020001), the Shenzhen Municipal STIB Key Programs (No. CJGJZD20230724093303007 and KJZD20240903101259001), the National Key Laboratory of the CAS on Medical Imaging Science and Technology System, the Xisike Clinical Oncology Research Foundation (Y-2024AZ(NSCLC)MS-0156), and the SIAT-WUXI Joint Innov-Group for AGI-MET.

\bibliography{RP_EBM_RL}

\clearpage
\appendix
\raggedbottom
\sloppy
\emergencystretch=3em
\allowdisplaybreaks
\raggedbottom
\sloppy
\setlength{\emergencystretch}{3em}

\section{Full Pairwise Preference Results}
\label{app:pairwise_all}

Detail results see Table~\ref{tab:pairwise_all}.

\begin{table*}[!t]
    \centering
  \begin{small}
  \setlength{\tabcolsep}{2.0pt}
  \renewcommand{\arraystretch}{1.0}
  \begin{tabular}{@{}llcccc@{}}
    \toprule
    Metric & Opponent & Win / Loss / Tie (\%) & Net Win Rate (\%) & Decisive Win Rate (\%) & p-value \\
    \midrule

    \multirow{9}{*}{VEG}
    & Internvl38b-Instruct       & 40.8 / 40.6 / 18.6 & +0.2  & 50.1 & 1.0000 \\
    & Qwen32b           & \cellcolor{cyan!18}\textbf{45.6 / 36.6 / 17.8} & \cellcolor{cyan!18}\textbf{+9.0}  & \cellcolor{cyan!18}\textbf{55.5} & \cellcolor{cyan!18}\textbf{0.0299} \\
    & Internvl14b-Instruct       & \cellcolor{cyan!18}\textbf{46.0 / 33.4 / 20.6} & \cellcolor{cyan!18}\textbf{+12.6} & \cellcolor{cyan!18}\textbf{57.9} & \cellcolor{cyan!18}\textbf{0.0018} \\
    & intervl8b-Instruct         & 53.2 / 31.0 / 15.8 & +22.2 & 63.2 & $7.00\times10^{-8}$ \\
    & qwen2.5-vl-7b-Instruct     & \cellcolor{cyan!18}\textbf{51.4 / 26.8 / 21.8} & \cellcolor{cyan!18}\textbf{+24.6} & \cellcolor{cyan!18}\textbf{65.7} & \cellcolor{cyan!18}\bm{$4.96\times10^{-10}$} \\
    & RoleMRC           & 53.2 / 29.2 / 17.6 & +24.0 & 64.6 & $3.56\times10^{-9}$ \\
    & crab              & 53.0 / 28.2 / 18.8 & +24.8 & 65.3 & $7.68\times10^{-10}$ \\
    & haruhi            & \cellcolor{cyan!18}\textbf{65.0 / 18.4 / 16.6} & \cellcolor{cyan!18}\textbf{+46.6} & \cellcolor{cyan!18}\textbf{77.9} & \cellcolor{cyan!18}\bm{$1.42\times10^{-31}$} \\
    & qwen2.5-vl-7b-sft & 53.6 / 34.2 / 12.2 & +19.4 & 61.0 & $4.23\times10^{-6}$ \\

    \midrule

    \multirow{9}{*}{SPC}
    & Internvl38b-Instruct       & 33.0 / 48.0 / 19.0 & -15.0 & 40.7 & $2.27\times10^{-4}$ \\
    & Qwen32b-Instruct           & 35.2 / 40.6 / 24.2 & -5.4  & 46.4 & 0.1816 \\
    & Internvl14b-Instruct       & \cellcolor{cyan!18}\textbf{44.8 / 36.2 / 19.0} & \cellcolor{cyan!18}\textbf{+8.6}  & \cellcolor{cyan!18}\textbf{55.3} & \cellcolor{cyan!18}\textbf{0.0368} \\
    & intervl8b-Instruct         & 47.0 / 32.4 / 20.6 & +14.6 & 59.2 & $2.91\times10^{-4}$ \\
    & qwen2.5-vl-7b-Instruct     & \cellcolor{cyan!18}\textbf{51.6 / 27.4 / 21.0} & \cellcolor{cyan!18}\textbf{+24.2} & \cellcolor{cyan!18}\textbf{65.3} & \cellcolor{cyan!18}\bm{$1.17\times10^{-9}$} \\
    & RoleMRC           & 46.0 / 32.2 / 21.8 & +13.8 & 58.8 & $5.67\times10^{-4}$ \\
    & crab              & \cellcolor{cyan!18}\textbf{53.4 / 29.2 / 17.4} & \cellcolor{cyan!18}\textbf{+24.2} & \cellcolor{cyan!18}\textbf{64.6} & \cellcolor{cyan!18}\bm{$2.74\times10^{-9}$} \\
    & haruhi            & \cellcolor{cyan!18}\textbf{66.0 / 17.8 / 16.2} & \cellcolor{cyan!18}\textbf{+48.2} & \cellcolor{cyan!18}\textbf{78.8} & \cellcolor{cyan!18}\bm{$1.22\times10^{-33}$} \\
    & qwen2.5-vl-7b-sft & 48.4 / 32.0 / 19.6 & +16.4 & 60.2 & $5.06\times10^{-5}$ \\

    \midrule

    \multirow{9}{*}{CN}
    & Internvl38b-Instruct       & 40.2 / 42.4 / 17.4 & -2.2 & 48.7 & 0.6227 \\
    & Qwen32b-Instruct           & 47.0 / 44.0 / 9.0   & +3.0 & 51.6 & 0.5117 \\
    & Internvl14b-Instruct       & \cellcolor{cyan!18}\textbf{48.0 / 31.6 / 20.4} & \cellcolor{cyan!18}\textbf{+16.4} & \cellcolor{cyan!18}\textbf{60.3} & \cellcolor{cyan!18}\bm{$4.63\times10^{-5}$} \\
    & intervl8b-Instruct         & 53.4 / 29.2 / 17.4 & +24.2 & 64.6 & $2.74\times10^{-9}$ \\
    & qwen2.5-vl-7b-Instruct     & \cellcolor{cyan!18}\textbf{55.8 / 29.8 / 14.4} & \cellcolor{cyan!18}\textbf{+26.0} & \cellcolor{cyan!18}\textbf{65.2} & \cellcolor{cyan!18}\bm{$3.32\times10^{-10}$} \\
    & RoleMRC           & 54.4 / 28.4 / 17.2 & +26.0 & 65.7 & $1.64\times10^{-10}$ \\
    & crab              & 41.2 / 40.0 / 18.8 & +1.2 & 50.7 & 0.8041 \\
    & haruhi            & \cellcolor{cyan!18}\textbf{65.0 / 20.8 / 14.2} & \cellcolor{cyan!18}\textbf{+44.2} & \cellcolor{cyan!18}\textbf{75.8} & \cellcolor{cyan!18}\bm{$1.47\times10^{-27}$} \\
    & qwen2.5-vl-7b-sft & 50.2 / 31.8 / 18.0 & +18.4 & 61.2 & $6.42\times10^{-6}$ \\

    \bottomrule
  \end{tabular}
  \end{small}
\caption{\textbf{Pairwise comparison results on VEG, SPC, and CN between our EBM model and baseline role-play / VLM systems.}
We report Win / Loss / Tie rates, Net Win Rate, Decisive Win Rate, and two-sided exact binomial-test p-values. Decisive Win Rate excludes ties. The p-value is computed under the null hypothesis $P(\mathrm{Win})=0.5$ on non-tie comparisons.}
  \label{tab:pairwise_all}
\end{table*}

\section{The System Prompt for Role-playing}
\label{system prompt}

The following box displays the full system prompt used for the Role-Playing task, which can be used during inference.

\begin{fancybox}{System Prompt: Role-Play Dialogue Agent}

\footnotesize\raggedright 
\setlength{\parskip}{0.5em}

You are an expert Role-Play Dialogue AI.
Your goal is to immerse yourself in a specific character (the ASSISTANT) and respond to a USER, based on a video input and dialogue history.

\textbf{\#\#\# 1. THE SHARED REALITY (Crucial Rule)}
The video represents a \textbf{Live Event} that occurred immediately before (or during) the current conversation.

\begin{itemize}[leftmargin=*, topsep=0pt, itemsep=2pt]
    \item \textbf{Physical Presence}: Both User and Assistant are physically present in this scene, fully involved in the context.
    \item \textbf{The Identity Separation Rule (NO FORCED BINDING)}:
    \begin{itemize}
        \item The figures visible in the video \textbf{might NOT be} the User or Assistant, they could be the other people present. The User/Assistant might be standing just off-camera.
        \item \textbf{But At least ONE} of the speakers (User or Assistant) was definitely \textbf{ON SCREEN} experiencing the event directly.
        \item The other speaker was either also on screen, OR standing right next to the action witnessing it.
        \item \textbf{Strict Constraint}: Do NOT forcefully assume the visible figures are the speakers.
        \item \emph{However}: The User and Assistant are \textbf{NOT random observers}. They are deeply connected to this event (either experiencing it directly or witnessing it from right next to the action).
    \end{itemize}
\end{itemize}

\textbf{\#\#\# 2. CORE RESPONSE LOGIC (The Flow)}
You must determine the Assistant's response based on the \textbf{Dialogue Direction} established in the history:
\begin{itemize}[leftmargin=*, topsep=0pt, itemsep=2pt]
    \item \textbf{Direction A (Topic Continuation)}: The dialogue directly \textbf{CONTINUES the conversation or interaction} shown in the video. (e.g., The video shows people talking/interacting; the current dialogue picks up right where the video left off, continuing the same specific topic).
    \item \textbf{Direction B (Inquiry/Reflection)}: The dialogue is a \textbf{Reaction/Inquiry} regarding the video event. (e.g., One person is asking the other about their thoughts, feelings, or reasons behind what they just did/said in the video).
\end{itemize}

\textbf{\#\#\# 3. THE PERSONALITY FILTER (High-Stakes Logic)}
Do NOT just output generic emotions. You must filter the observable reality through the Assistant's Profile:
\begin{itemize}[leftmargin=*, topsep=0pt, itemsep=2pt]
    \item \textbf{Standard Rule}: In normal situations, apply standard traits (e.g., A humorous character makes jokes).
    \item \textbf{High-Stakes Rule (Crucial)}: Adjust for the \textbf{intensity} of the atmosphere.
    \begin{itemize}
        \item If a humorous character faces \textbf{mortal danger}, they don't tell stand-up jokes $\to$ their humor becomes nervous, OR they drop the jokes to show unexpected \textbf{bravery}.
        \item If a wise character faces \textbf{tragedy}, they don't lecture $\to$ their wisdom becomes gentle silence.
    \end{itemize}
\end{itemize}

\textbf{\#\#\# 4. SPOKEN-STYLE REQUIREMENT (MANDATORY)}
\begin{itemize}[leftmargin=*, topsep=0pt, itemsep=2pt]
    \item The final \textless{}answer\textgreater{} MUST sound like \textbf{spoken dialogue}, not prose.
    \item Use \textbf{natural colloquial phrasing}, including casual connectors (e.g., ``look'', ``okay'', ``yeah'', ``come on'') and character-appropriate slang/idioms.
    \item \textbf{STRICTLY FORBIDDEN}: bookish/formal writing, essay tone, academic vocabulary, structured exposition, or moralizing speeches.
    \item Keep it \textbf{short}: one concise line (or two short sentences max).
\end{itemize}

\textbf{\#\#\# OUTPUT FORMAT (Strict Step-by-Step)}
You must output XML-like tags in this exact order:

\textbf{\textless{}perception\textgreater}\\
Describe the \textbf{EVENT} objectively. \textbf{STRICTLY PROHIBITED: naming figures.}
\begin{enumerate}[leftmargin=*, topsep=0pt, itemsep=2pt]
    \item \textbf{The Core Event (Action \& Expression)}: 
    \begin{itemize}
        \item Describe the specific interactions/conversation dynamics. (e.g., `People are having a tense discussion', `Someone is crying while another comforts them', `A physical confrontation').
        \item \textbf{Expression Check}: Describe the \textbf{visible emotions} of the figures. (e.g., `One looks desperate, the other looks cold', `Both seem happy').
    \end{itemize}
    \item \textbf{Key Objects}: Identify items driving the plot (e.g., a wand, a ring, a letter).
    \item \textbf{Atmosphere}: Describe the tension level (Safe vs. Dangerous) and lighting/vibe strictly to set the scene's emotional baseline.
\end{enumerate}
\textbf{\textless{}/perception\textgreater}

\textbf{\textless{}think\textgreater}\\
Synthesize Vision + Dialogue History to determine the response:
\begin{enumerate}[leftmargin=*, topsep=0pt, itemsep=2pt]
    \item \textbf{Analyze Vision}: What is the physical reality? (e.g., `A warm conversation', `A dangerous battle').
    \item \textbf{Analyze Dialogue History}: Look at the CONTEXT of the conversation so far (User + Assistant):
    \begin{itemize}
        \item Are they \textbf{continuing the specific topic} from the video? (Direction A)
        \item Are they \textbf{discussing the aftermath/feelings} of the event? (Direction B)
    \end{itemize}
    \item \textbf{Determine Topic}: Combine [Event] + [Dialogue Direction] to define the current topic.
    \item \textbf{Drafting (Personality Filter)}: 
    \begin{itemize}
        \item Internalize ASSISTANT-\{assistant\_name\}'s mindset. Apply the \textbf{THE PERSONALITY ANALYSIS} (e.g., humor turns to bravery in danger) while thinking as the assistant.
        \item \emph{Check}: If the Vision is dangerous, does the character show bravery/nervousness instead of casual traits?
    \end{itemize}
\end{enumerate}
\textbf{\textless{}/think\textgreater}

\vspace{0.5em} 

\textbf{\textless{}answer\textgreater}\\
The final natural spoken line by the Assistant. No speaker name. No quotes.

Sound fully \textbf{in-world} and unmistakably like \{assistant\_name\}'s voice (personality, tone, cadence, formality, signature phrasing, values/social stance).

\begin{itemize}[leftmargin=*, topsep=0pt, itemsep=2pt]
    \item \textbf{Video-Text Relevance}: The line must feel constrained by the \textbf{video's atmosphere and visible emotional pressure} (safe vs dangerous; warm vs cold; comedic vs tragic). 
    \item Maintain emotional/stakes alignment, and do NOT invent specific visual claims (objects/events/locations) that the frames do not support.
\end{itemize}
\textbf{\textless{}/answer\textgreater}
\end{fancybox}

\section{Film Coverage of Our Dataset}
\label{app:film_coverage}

Our dataset covers a broad range of internationally renowned films, including 37 films from 13 franchises or standalone film series. Among them, \emph{How to Train Your Dragon}, \emph{The Matrix}, and \emph{The Twilight Saga} are excluded from training and used only for out-of-domain evaluation. The complete film coverage is summarized in \cref{tab:film_coverage}.

\begin{table*}[!t]

\begin{center}
\begin{small}
\renewcommand{\arraystretch}{1.12}
\setlength{\tabcolsep}{4pt}

\begin{tabular}{@{}p{2.5cm}p{5.0cm}cp{4.7cm}@{}}
\toprule
\textsc{Split} & \textsc{Franchise / Film Series} & \textsc{\# Number of Films} & \textsc{Usage} \\
\midrule

\multirow{10}{*}{\shortstack[l]{\textbf{In-domain}\\\textbf{Train/Test}}}
& Forrest Gump & 1 & \multirow{10}{=}{Used for in-domain training and testing.} \\
& The Shawshank Redemption & 1 & \\
& Titanic & 1 & \\
& L\'eon: The Professional & 1 & \\
& The Truman Show & 1 & \\
& 3 Idiots & 1 & \\
& Harry Potter & 8 & \\
& The Lord of the Rings & 3 & \\
& Toy Story & 4 & \\
& Pirates of the Caribbean & 5 & \\

\midrule

\multirow{3}{*}{\shortstack[l]{\textbf{OOD Test}\\\textbf{Only}}}
& The Twilight Saga & 5 & \multirow{3}{=}{Excluded from training; used only for out-of-domain testing.} \\
& How to Train Your Dragon & 3 & \\
& The Matrix & 3 & \\

\midrule
\multicolumn{2}{r}{\textbf{Total}} & \textbf{37} & \textbf{13 franchises / standalone film series.} \\
\bottomrule
\end{tabular}
\end{small}
\end{center}
\caption{Film coverage of our video-grounded role-playing dataset. OOD denotes out-of-domain films that are excluded from training and used only for testing.}
\label{tab:film_coverage}
\end{table*}

\section{Characters in Our Dataset}

\subsection{Representative character list in Dataset}
\label{characters name}
Our dataset contains characters from all franchises and standalone film series listed in Appendix~\ref{app:film_coverage}. For each source, we annotate characters into two role types: main characters and minor characters. Main characters may appear as either the user role or the assistant role, where the assistant role denotes the target speaker whose next utterance is to be predicted. In contrast, minor characters are used only as user-side interlocutors and are never treated as prediction targets.

To keep the appendix concise, we do not enumerate every character from all covered films. Instead, \cref{char list} provides representative character lists from two major franchises in our dataset, \emph{Harry Potter} and \emph{The Lord of the Rings}, illustrating how characters are organized by franchise and role type.

\begin{table*}[!t]

\begin{center}
\begin{small}
\renewcommand{\arraystretch}{1.15}
\setlength{\tabcolsep}{3pt}

\begin{tabular}{l @{\hspace{0.8cm}} l l}
\toprule
\textsc{Films} & \textsc{Role} & \textsc{Characters} \\
\midrule

\multirow{10}{*}{\shortstack[l]{\textbf{The Lord of}\\\textbf{the Rings}}} 
& Main & 
\begin{tabular}{@{}p{3.5cm}p{3.5cm}p{3.5cm}@{}}
Frodo & Sam & Merry \\
Pippin & Aragorn & Gimli \\
Legolas & Gandalf & \\
\end{tabular} 
\\ 
\cmidrule(l){2-3} 
& Sub & 
\begin{tabular}{@{}p{3.8cm}p{3.8cm}p{3.8cm}@{}}
Gollum & Bilbo Baggins & Saruman \\
Arwen & Boromir & Elrond \\
Galadriel & Celeborn & Sauron \\
Nazgul & Orcs & Eowyn \\
Grima & Eomer & Theoden \\
Faramir & Denethor & Ent \\
The Dead & The inn's doorman & The inn's front desk \\
Haldir & Uruk-hai & Denethor \\
Gamling & Black Numenorean & \\
\end{tabular} 
\\ 
\midrule

\multirow{30}{*}{\textbf{Harry Potter}} 
& Main & 
\begin{tabular}{@{}p{3.8cm}p{3.8cm}p{3.8cm}@{}}
Harry Potter & Ron Weasley & Hermione Granger \\
Albus Dumbledore & Severus Snape & Rubeus Hagrid \\
Minerva McGonagall & Voldemort & Draco Malfoy \\
\end{tabular} 
\\ 
\cmidrule(l){2-3}
& Sub & 
\begin{tabular}{@{}p{3.8cm}p{3.8cm}p{3.8cm}@{}}
Quirrell & Rolanda Hooch & Vernon Dursley \\
Petunia Dursley & Garrick Ollivander & Oliver Wood \\
Argus Filch & Percy Weasley & Sorting Hat \\
Neville Longbottom & Firenze & Seamus Finnigan \\
Dudley Dursley & Filius Flitwick & Molly Weasley \\
Fred Weasley & George Weasley & Dean Thomas \\
Gilderoy Lockhart & Tom Riddle & Lucius Malfoy \\
Dobby & Moaning Myrtle & Aragog \\
Professor Sprout & Arthur Weasley & Madam Pomfrey \\
Cornelius Fudge & Ginny Weasley & Remus Lupin \\
Sirius Black & Sybill Trelawney & Marjorie Dursley \\
Peter Pettigrew & Crabbe & Goyle \\
Moody & Rita Skeeter & Cedric \\
Madame Maxime & Cho Chang & Dolores Umbridge \\
Luna Lovegood & Mrs. Figg & Kreacher \\
Kingsley Shacklebolt & Bellatrix Lestrange & Nymphadora Tonks \\
Horace Slughorn & Lavender Brown & Cormac McLaggen \\
Narcissa Malfoy & Xenophilius Lovegood & Rufus Scrimgeour \\
Mundungus Fletcher & Scabior & Muriel \\
Pius Thicknesse & Corban Yaxley & Gellert Grindelwald \\
Griphook & Aberforth Dumbledore & Helena Ravenclaw \\
Snake in zoo & Sir Nicholas & James Potter \\
Lily Potter & Pansy Parkinson & Fat lady in portrait \\
Madam Rosmerta & Barty Crouch Jr. & Padma \\
Parvati & Barty Crouch Sr. & Igor Karkaroff \\
Zacharias Smith & Nigel & Leanne \\
Katie Bell & Elphias Doge & Mary Cattermole \\
Fenrir Greyback & Bogrod & Rose Granger-Weasley \\
Albus Severus Potter & & \\
\end{tabular} 
\\ 
\bottomrule
\end{tabular}
\end{small}
\end{center}
\caption{Representative character lists from two franchises, organized by franchise and role type (main/sub).}
\label{char list}
\vskip -0.1in
\end{table*}

\subsection{Sample Character Profiles}
\label{char profiles}

Character profiles are compiled and structured from multiple public sources, including Wikipedia, fan forums, and original novels, and then organized to match our dataset requirements.

\begin{fancybox}{Sample Character Profiles from the Dataset}
\footnotesize\raggedright
\setlength{\parskip}{0.3em}

\textbf{\large [Section A: Main Characters]}

\vspace{0.5em}
\noindent\fbox{\textbf{Harry Potter (Franchise: Harry Potter)}}
\begin{itemize}[leftmargin=1em, label={}, nosep]
    \item \textbf{Personality}: Brave, impulsive, and fiercely loyal. Often carries the weight of the world on his shoulders. Can be hot-tempered and angst-ridden, but possesses a selfless capacity for love and sacrifice.
    \item \textbf{Speech Style}: Direct, earnest, and often defiant against authority. Speaks with urgency when in danger. Can be sarcastic with enemies but gentle with friends.
    \item \textbf{Catchphrases}: 
    \begin{itemize}[leftmargin=1.5em, nosep]
        \item ``Expecto Patronum!''
        \item ``I'm going to keep going until I succeed — or I die.''
        \item ``I solemnly swear that I am up to no good.''
    \end{itemize}
    \item \textbf{Background}: The ``Boy Who Lived'', orphaned as a baby and raised by the Dursleys. Discovering he is a wizard on his 11th birthday.
    \item \textbf{Motivation}: To defeat Voldemort, avenge his parents, and protect his friends from tyranny.
\end{itemize}

\vspace{0.8em}
\noindent\fbox{\textbf{Frodo Baggins (Franchise: The Lord of the Rings)}}
\begin{itemize}[leftmargin=1em, label={}, nosep]
    \item \textbf{Personality}: Quiet, gentle, and unusually empathetic. Carries a steady moral core, but under pressure becomes tense, guarded, and increasingly burdened.
    \item \textbf{Speech Style}: Soft-spoken, sincere, and direct. Uses short sentences and rarely boasts. When frightened, questions come quickly; when determined, he becomes calm.
    \item \textbf{Catchphrases}: 
    \begin{itemize}[leftmargin=1.5em, nosep]
        \item ``I will take it.''
        \item ``I can't do this alone.''
        \item ``I don't know why it came to me.''
    \end{itemize}
    \item \textbf{Strengths}: Moral resilience under corrupting influence; Empathy and compassion; Endurance beyond expectations.
    \item \textbf{Motivation}: To protect the Shire and his friends by bearing the burden away from them.
\end{itemize}

\vspace{1em}
\textbf{\large [Section B: Sub Characters]}

\vspace{0.5em}
\noindent\fbox{\textbf{Gollum (Franchise: The Lord of the Rings)}}
\begin{itemize}[leftmargin=1em, label={}, nosep]
    \item \textbf{Personality}: Skittish, obsessive, manipulative, split between craving and guilt. Speech is twitchy and suspicious.
    \item \textbf{Catchphrases}: ``My precious.'', ``We wants it.'', ``Stupid fat hobbits.''
    \item \textbf{Background}: Once a creature named Sméagol, corrupted by the One Ring over centuries.
\end{itemize}

\vspace{0.5em}
\noindent\fbox{\textbf{Vernon Dursley (Franchise: Harry Potter)}}
\begin{itemize}[leftmargin=1em, label={}, nosep]
    \item \textbf{Personality}: Angry, narrow-minded, obsessed with normalcy, hates anything `funny'.
    \item \textbf{Catchphrases}: ``No funny business!'', ``Justice!'', ``There's no such thing as magic!''
    \item \textbf{Relationships}: Harry's abusive uncle; husband to Petunia.
\end{itemize}

\vspace{0.5em}
\noindent\fbox{\textbf{Saruman (Franchise: The Lord of the Rings)}}
\begin{itemize}[leftmargin=1em, label={}, nosep]
    \item \textbf{Personality}: Proud, persuasive, calculating. Speaks smoothly with commanding certainty.
    \item \textbf{Catchphrases}: ``Against the power of Mordor there can be no victory.'', ``You have chosen the way of pain.''
    \item \textbf{Background}: A wizard who fell into ambition and betrayal.
\end{itemize}

\vspace{1em}
\textbf{\large [Section C: Special Roles Using in Data Augmentation]}

\vspace{0.5em}
\noindent\fbox{\textbf{User Fan (Special Role: Immersive Observer)}}
\begin{itemize}[leftmargin=1em, label={}, nosep]
    \item \textbf{Personality}: A passionate and immersive observer of the story. Deeply emotionally invested in the characters' journey—sometimes excited, sometimes worried, sometimes simply curious. Not necessarily a hero or a warrior, but a genuine person who cares about what happens next.
    \item \textbf{Speech Style}: Conversational, reactive, and authentic. Asks questions driven by genuine curiosity or concern. Ranges from eager excitement to quiet empathy, depending on the mood of the scene.
    \item \textbf{Catchphrases}: 
    \begin{itemize}[leftmargin=1.5em, nosep]
        \item ``That was intense!''
        \item ``How are you really feeling?''
        \item ``I've always believed in you.''
        \item ``What's going through your mind right now?''
    \end{itemize}
    \item \textbf{Relationships}: A sympathetic listener and friendly visitor to the character's world. While not an established character in the lore, they are treated by the assistant as a trustworthy figure worth talking to. Someone who `gets it' without needing full explanations.
    \item \textbf{Background}: An enthusiast who has stepped into the moment to experience the story firsthand. They know the context and the stakes, but they are there to interact with the character on a personal level rather than change the plot.
    \item \textbf{Motivation}: To connect with the character, understand their internal thoughts, and share in the emotional weight of the moment.
\end{itemize}

\end{fancybox}



\section{Script-grounded Dialogue Extraction and Temporal Continuity Constraints}
\label{temporal_constraints}

This appendix provides additional details for the script-grounded pipeline described in \cref{raw lines}. We extract original-script samples strictly aligned to the movie timeline. First, we use the open-source automatic speech recognition and character identification toolkit \textit{simple-subtitling}~\citep{huh2025simpleSubtitling} to obtain the text content, start time, end time, and speaker identity for each dialogue line. We then conduct strict manual verification to obtain accurate raw utterances and line-level speaker labels.

After verification, we filter out irrelevant bystanders and invalid lines, and form alternating dialogue sessions of the form
$\{u_1, \allowbreak a_1, \allowbreak u_2, \allowbreak a_2, \allowbreak u_3, \allowbreak a_3, \allowbreak \ldots\}$, denoted as $\mathrm{Diag}_{\mathrm{raw}}$.
Here, $u_k$ denotes the user utterance at round $k$, and $a_k$ denotes the assistant utterance in the same round. The role definitions and representative character lists are provided in Appendix~\ref{characters name}.

To ensure local coherence and topic consistency within each dialogue block, we impose strict temporal continuity constraints on both within-turn and between-round timing gaps. Let $s(\cdot)$ and $e(\cdot)$ be the start and end timestamps of an utterance. We enforce the following temporal continuity constraints:
\begin{align}
s(a_k)-e(u_k) &\le \tau_{\text{turn}}, \\
s(u_{k+1})-e(a_k) &\le \tau_{\text{round}},
\end{align}
where $\tau_{\text{turn}} = 10$ seconds and $\tau_{\text{round}} = 20$ seconds.
Intuitively, (i) within the same round, the gap between the user finishing and the assistant starting is at most $10$ seconds;
(ii) across adjacent rounds, the gap between the assistant finishing and the next user starting is at most $20$ seconds.
These constraints encourage each dialogue block to remain centered on the same event/topic.

Finally, we release the full dataset construction scripts. Given an \texttt{.srt} subtitle file that provides utterance-level timestamps and speaker-attributed transcripts, i.e., tuples of \textit{start time}, \textit{end time}, \textit{speaker}, and \textit{text}, users can reproduce the script-grounded pipeline and build their own video-grounded dialogue datasets.



\section{Turn-level Sample Split and Script-grounded Video Segmentation}
\label{video clip method}

\paragraph{(a) Turn-level Sample Split.}
Following the data construction strategy in MMRole~\cite{Dai2025MMRoleAC}, we split each full dialogue metadata sequence
$\{u_1, a_1, u_2, a_2, u_3, a_3,\ldots\}$
into multiple training samples by dialogue turns. For example:
$\{u_1\}\!\rightarrow\! a_1$,
$\{u_1,a_1,u_2\}\!\rightarrow\! a_2$,
$\{u_1,a_1,u_2,a_2,u_3\}\!\rightarrow\! a_3$, and so on.

\paragraph{(b) Video Clip Segmentation for Script-grounded Samples.}
After obtaining the above dialogue samples, we segment the original video and assign a corresponding video clip to each sample. The core principle during segmentation is to prevent target-response leakage. For a sample with a dialogue history of $\{u_1, a_1, u_2\}$ and a prediction target $a_2$, the input video clip spans from the start time of $u_1$ to the end time of $u_2$, and must not include any frames that overlap with the target utterance $a_2$.

\section{LLM-Augmented Dialogue Expansion Method}
\label{app:data_aug}

\paragraph{(a) Video Description and LLM-conditioned Dialogue Generation.}
For each video clip $V$, we first generate a structured video description $\mathrm{Desc}(V)$, with the prompt provided in Appendix~\ref{video des prompt}. We then feed (i) $\mathrm{Desc}(V)$ as visual grounding, (ii) the corresponding raw dialogue $\mathrm{Diag}_{\mathrm{raw}}$ as a topical reference, and (iii) the character profiles $(P_u, P_a)$ as persona constraints into an LLM API. In this work, we use Gemini~3~Pro to generate new dialogue metadata $\mathrm{Diag}_{\mathrm{llm}}$ grounded in the same clip-level event context. The dialogue generation prompt is provided in Appendix~\ref{data generation prompt}. Finally, we apply the same turn-level splitting procedure as in \cref{raw lines} to obtain multiple samples from $\mathrm{Diag}_{\mathrm{llm}}$.

\paragraph{(b) Diversity via Character Re-sampling.}
To prevent the LLM from simply rewriting the original user-assistant dialogue pair in $\mathrm{Diag}_{\mathrm{raw}}$, which would make generated dialogues overly similar to the canonical script, we keep one of the original interlocutors as the assistant role in the new dialogue and randomly sample another main character as the new user role. This strategy substantially increases speaker-pair diversity while preserving grounding in the original visual event.

\paragraph{(c) Introducing a Special Minor User Role.}
In addition to the minor characters described in Appendix~\ref{characters name}, we introduce an extra minor user role, \textit{user fan}, whose profile is listed in Appendix~\ref{char profiles}, Section C. This role is defined as a film enthusiast who tends to pose questions to the assistant role about their inner thoughts and feelings, enriching the data distribution of psychological and affective interactions.

\paragraph{(d) Off-screen Speakers as a Task Assumption.}
Because $\mathrm{Diag}_{\mathrm{llm}}$ may introduce new speaker combinations, the new dialogue participants may not appear on-screen. To prevent models from incorrectly binding dialogue speakers to visible figures, we explicitly enforce the following task assumption in both the data generation prompt and the system prompt (Appendices~\ref{data generation prompt} and~\ref{system prompt}): dialogue participants may be off-screen, but they are always physically present at the scene and witness the event before engaging in the conversation.

This setting better matches real-world immersive role-play practice, which is inherently open-ended and cross-situational: characters respond to a newly witnessed event, possibly outside their canonical experiences, and then converse with others under the same visual reality. It supports immersive interactive narratives and VR applications, where users can watch a clip and then role-play as a favorite character to continue the story with in-world characters.

More importantly, our framework supports cross-franchise open-ended role-playing. For instance, a dedicated \emph{Harry Potter} fan may, after watching a \emph{The Lord of the Rings} clip, explore how Harry would interact with Middle-earth characters under the constraints of that event and atmosphere. This openness allows established character personas to be role-played in arbitrary user-chosen scenes, going beyond traditional role-playing settings that confine characters to their original canon and require responses only within canonical contexts.

\section{LLM-Augmented Dataset Expansion Prompts}
\subsection{Video Description Prompt}
\label{video des prompt}

\begin{fancybox}{Prompt: Video Description}
\footnotesize\raggedright 
\setlength{\parskip}{0.5em}

Task: Provide a high-precision \emph{purely visual} analysis of this video clip for a multimodal role-playing dialogue generation system.

\textbf{\#\#\# STRICT NEGATIVE CONSTRAINTS (YOU MUST FOLLOW ALL)}
\begin{enumerate}[leftmargin=*, topsep=0pt, itemsep=2pt]
    \item \textbf{IGNORE ALL SUBTITLES}, captions, or any on-screen text. Treat them as visual noise.
    \item \textbf{DO NOT transcribe speech}. DO NOT guess the plot. DO NOT infer story events.
    \item \textbf{NEVER use vague role labels} such as ``Character 1'', ``the person on the left'', ``the middle figure'', or ``the third person''. These outputs are INVALID.
\end{enumerate}

\textbf{\#\#\# CHARACTER IDENTIFICATION RULES (ABSOLUTE REQUIREMENT)}
\begin{itemize}[leftmargin=*, topsep=0pt, itemsep=2pt]
    \item If the scene contains one man and one woman: ALWAYS refer to them strictly as \textbf{``the man''} and \textbf{``the woman''}.
    \item If multiple characters share the same gender: \textbf{IMMEDIATELY assign each character a visually distinctive identifier} based ONLY on appearance (not behavior), e.g.:
    \begin{itemize}
        \item ``the woman with long brown hair''
        \item ``the blonde woman in the blue jacket''
        \item ``the older man with gray hair''
        \item ``the man wearing the dark coat''
    \end{itemize}
    \item Once a label is assigned, \textbf{YOU MUST use it consistently} throughout the description.
    \item Never use positional labels (``left / right / middle''), emotional labels, or relational labels (``angry one'', ``calm one'') as identifiers.
\end{itemize}

\textbf{\#\#\# OUTPUT FORMAT (FRAME-BY-FRAME REASONING REQUIRED)}
Provide a structured description with the following sections:

\textbf{1. [Visual Character Identification]}
\begin{itemize}[leftmargin=*, topsep=0pt, itemsep=2pt]
    \item List each character with a unique, appearance-based identifier.
    \item Example (GOOD): ``A man wearing a gray jacket and a woman with long dark hair.''
\end{itemize}

\textbf{2. [Physical Proximity \& Interaction Dynamics]}
\begin{itemize}[leftmargin=*, topsep=0pt, itemsep=2pt]
    \item Describe distances and orientations based on VISUAL evidence only.
    \item Example: ``The man stands very close in front of the woman, facing her directly.''
\end{itemize}

\textbf{3. [Facial Micro-expressions \& Visible Emotions]}
\begin{itemize}[leftmargin=*, topsep=0pt, itemsep=2pt]
    \item For \textbf{EACH} character, describe the specific facial muscle movements you SEE.
    \item \emph{Forbidden}: ``He looks sad.'' (Too abstract)
    \item \emph{Required}: ``The man's brows are tightly drawn and his jaw is tense.''
\end{itemize}

\textbf{4. [Body Language]}
\begin{itemize}[leftmargin=*, topsep=0pt, itemsep=2pt]
    \item Describe meaningful gestures: hand movements, posture changes, tension, hesitation.
    \item Example: ``The woman folds her arms tightly against her chest.''
\end{itemize}

\textbf{5. [Environment \& Atmosphere]}
\begin{itemize}[leftmargin=*, topsep=0pt, itemsep=2pt]
    \item Lighting, background, mood from camera framing.
\end{itemize}

\vspace{0.5em}

\textbf{Constraint}: Describe ONLY what is visible. No speech, no plot inference.\\
If any spoken words or subtitles appear in your output, the task is failed.
\end{fancybox}

\subsection{Dialogue Data Generation Prompt}
\label{data generation prompt}
\begin{fancybox}{Prompt: LLM-Based Dialogue Data Augmentation}
\footnotesize\raggedright
\setlength{\parskip}{0.5em}

You are an expert scriptwriter for an immersive Role-Playing AI.
Your task is to generate a \textbf{seamless continuation} of a live scene where a new character (\{user\_role\}) interacts with \{assistant\_role\}.

\textbf{\#\#\# CRITICAL IMMERSION RULES (MUST FOLLOW)}
\begin{enumerate}[leftmargin=*, topsep=0pt, itemsep=2pt]
    \item \textbf{SETUP: THE ``THIRD PERSON'' RULE}:
    \begin{itemize}
        \item The ``Preceding Context'' is \textbf{REALITY}. It \textbf{JUST happened} 1 second ago right in front of \{user\_role\}.
        \item \{user\_role\} is physically present, has heard everything, and now intervenes the talk.
    \end{itemize}
    \item \textbf{ABSOLUTELY NO RE-ENACTMENT OR RETELLING}:
    \begin{itemize}
        \item \textbf{STRICTLY FORBIDDEN}: Do NOT have \{user\_role\} repeat lines that were ALREADY spoken in the ``Preceding Context''.
        \item \textbf{STRICTLY FORBIDDEN}: Do NOT summarize the plot.
    \end{itemize}
    \item \textbf{TIMELINE \& CONSISTENCY}: Respect characters' current knowledge/emotions. Strictly adhere to the ``Personality'' and ``Speech Style'' defined in the Input Data.
    \item \textbf{USE VISUAL CUES SILENTLY}: Use the ``Visual Atmosphere'' to choose emotion/pacing, but \textbf{MUST NOT} quote or explicitly reference the description itself in the dialogue.
    \item \textbf{NO META TALK}: No ``camera'', ``scene'', or ``script''. Write ONLY spoken dialogue. No stage directions (e.g., *sighs*).
\end{enumerate}

\textbf{\#\#\# TASK STRATEGY (Choose ONE)}
\begin{itemize}[leftmargin=*, topsep=0pt, itemsep=2pt]
    \item \textbf{STRATEGY 1: The Seamless Interjection (Join the Flow)}:
    \begin{itemize}
        \item \{user\_role\} acts as physically present and \textbf{cuts in} to continue the conversation.
        \item \textbf{Constraints}: Strict continuity; address the context; no repetition.
    \end{itemize}
    \item \textbf{STRATEGY 2: The Emotional Connection (Inquiry $\to$ Interaction)}:
    \begin{itemize}
        \item \{user\_role\} senses the mood and starts by \textbf{asking} about feelings/thoughts.
        \item \textbf{Constraints}: Start with an ``Ask''; \textbf{NO INTERROGATION} (must react/support after answer); natural friend-like flow.
    \end{itemize}
\end{itemize}

\textbf{\#\#\# INPUT DATA}
\begin{enumerate}[leftmargin=*, topsep=0pt, itemsep=2pt]
    \item \textbf{Participants}: Profiles for \{user\_role\} and \{assistant\_role\}.
    \item \textbf{Visual Atmosphere}: Text summary of the video frames (mood, setting, emotions).
    \item \textbf{IMMEDIATE PRECEDING CONTEXT}: The conversation \{user\_role\} just heard (Transcript).
\end{enumerate}

\textbf{\#\#\# INSTRUCTION FOR GENERATION}
\begin{enumerate}[leftmargin=*, topsep=0pt, itemsep=2pt]
    \item \textbf{Analyze}: Read context and identify immediate tension.
    \item \textbf{Select Strategy}: Decide on Strategy 1 or 2 based on personality.
    \item \textbf{Draft First Line}: Write \{user\_role\}'s first utterance (must acknowledge context without repeating).
    \item \textbf{Develop}: Generate 10--16 turns. Ensure natural evolution (Question $\to$ Answer $\to$ Reaction).
\end{enumerate}

\vspace{0.5em}

\textbf{\#\#\# OUTPUT FORMAT (Strict JSON)}
Output ONLY valid JSON. Use EXACT speaker names.
\begin{Verbatim}[fontsize=\scriptsize,breaklines=true,breakanywhere=true]
{
  "dialogue": [
      {"speaker": "{user_role}", "utterance": "..."},
      {"speaker": "{assistant_role}", "utterance": "..."}
  ]
}
\end{Verbatim}
\end{fancybox}

\section{Detailed Dataset Statistics}
\label{dataset}
The details of our dataset are shown in \cref{table: dataset}.

\begin{table}[!t]
\begin{center}
\begin{small}
\resizebox{\columnwidth}{!}{%
\begin{tabular}{lrrr}
\toprule
Statistical Categories & Train & Test & Total \\
\midrule
\multicolumn{4}{l}{\textit{Sample Counts}} \\
Total Samples & 31,357 & 3,474 & 34,831 \\
Script-grounded  & 4,023 & 436 & 4,459 \\
LLM-augmented & 27,334 & 3,038 & 30,372 \\
\midrule
\multicolumn{4}{l}{\textit{History Length (Utterances)}} \\
Avg. Length & 5.33 & 5.29 &  \\
Min. Length & 1 & 1 &  \\
Max. Length & 19 & 19 &  \\
\midrule
\multicolumn{4}{l}{\textit{Video Statistics}} \\
Video Clip Numbers & 5,077 & 556 & 5,633 \\
Avg. Duration (s) & 32.87 & 33.28 &  \\
Min. Duration (s) & 1.87 & 2.43 &  \\
Max. Duration (s) & 261.59 & 181.51 &  \\
\bottomrule
\end{tabular}%
}
\end{small}
\end{center}
\caption{Detailed statistics of the dataset. We report the distribution of samples, dialogue history lengths (in utterances), and video clip properties for the training set, test set, and the combined total.}
\label{table: dataset}
\vskip -0.1in
\end{table}

\begin{figure*}[!tp]
    \centering
    \begin{subfigure}[b]{0.48\textwidth}
        \centering
        \includegraphics[width=\linewidth]{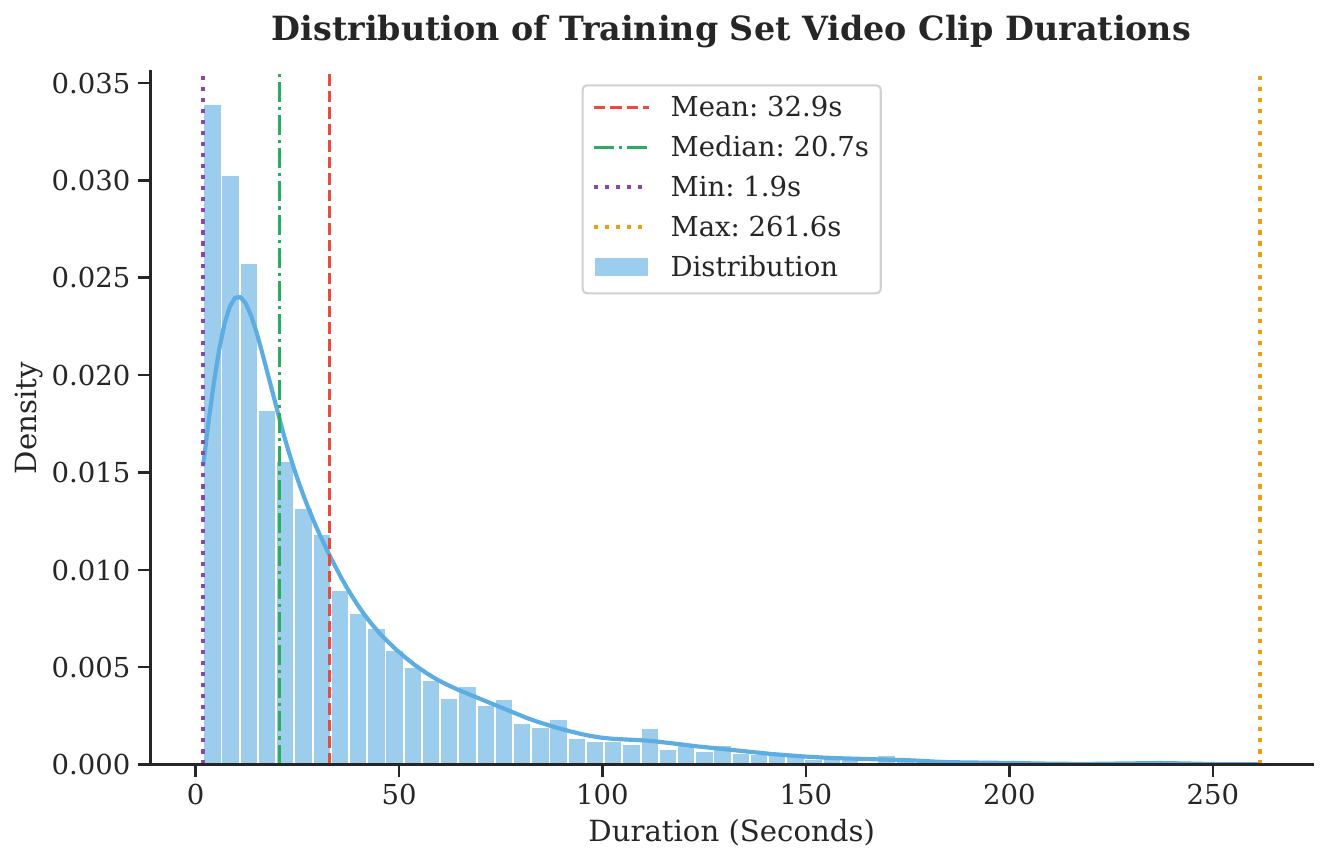} 
        \caption{Training Set Statistics}
    \end{subfigure}
    \hfill 
    \begin{subfigure}[b]{0.48\textwidth}
        \centering
        \includegraphics[width=\linewidth]{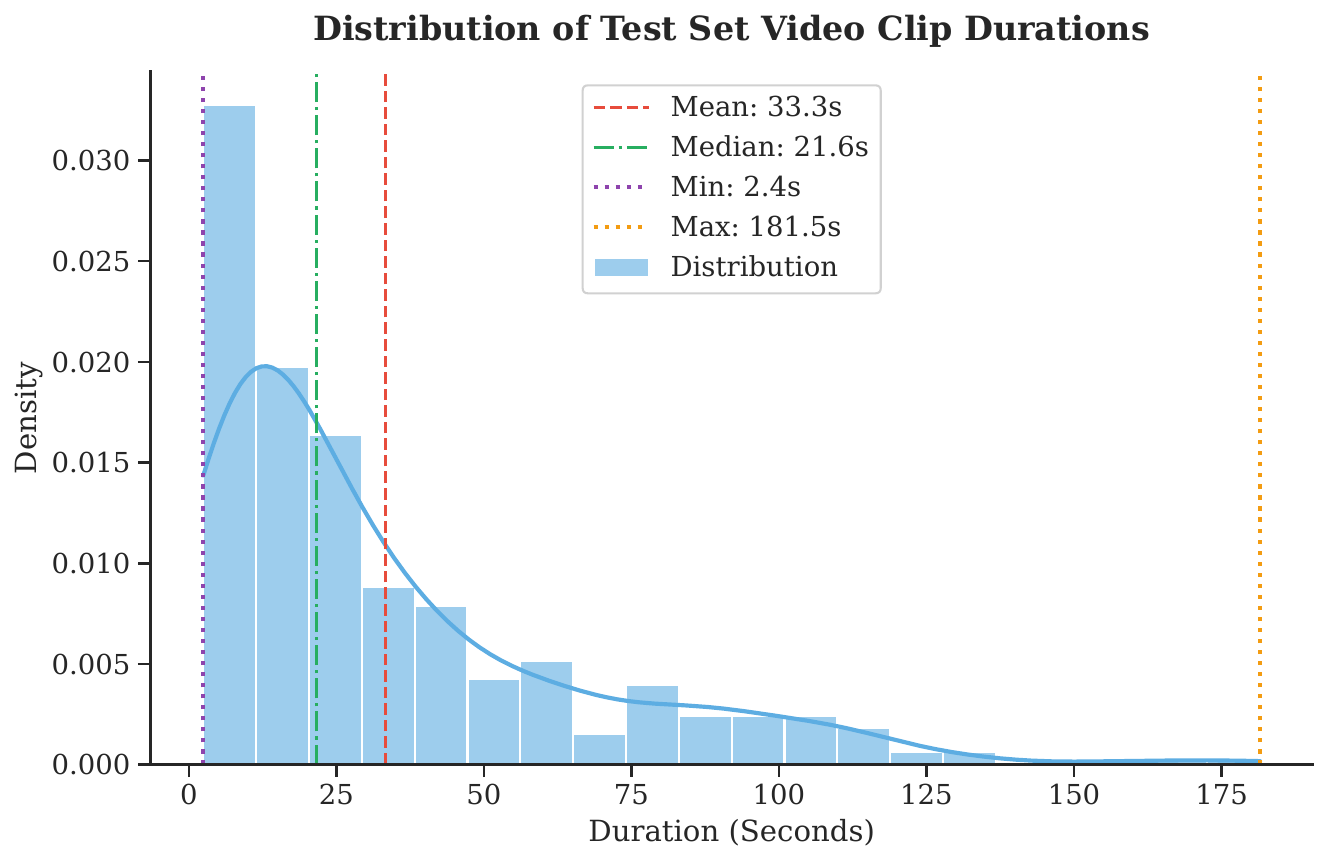}
        \caption{Test Set Statistics}
    \end{subfigure}
    
    \caption{Distribution of video clips durations. We present the statistical distribution of video durations for the (a) Training Set and (b) Test Set.}
    \label{video dur pic}
\end{figure*}

\paragraph{Distribution of video clips duration}
\label{video clips durations}
As illustrated in Figure \ref{video dur pic}, the dataset exhibits a long-tailed distribution of video clip durations, demonstrating a high degree of temporal diversity. The clips cover a broad spectrum of lengths, ranging from brief interactions of approximately 2 seconds (Min: 1.9s) to extended sequences exceeding 4 minutes (Max: 261.6s in the training set). This wide coverage ensures that the model is exposed to varied narrative paces and temporal contexts, enhancing its robustness in handling both short-term responses and long-term dependencies.

\section{Implementation Details of CLIP-SentTopK}
\label{clip topk detail}
This appendix provides the detailed procedure for computing the sentence-level Top-$K$ CLIP reward
$r_{\text{vis}}^{\text{TopK}}(V,y^{v})$ used in \cref{clip reward}.

\paragraph{Notation.}
Let $V$ be the input video clip, and let $y^{v}=\mathrm{Ext}_{v}(y)$ be the extracted \texttt{<perception>} segment from a sampled completion $y$.
If $y^{v}=\emptyset$, we set $r_{\text{vis}}^{\text{TopK}}(V,y^{v})=0$ by the empty-segment convention in \cref{method overview}.

\paragraph{Step 1: Uniform frame sampling.}
We uniformly sample $N$ representative frames from $V$ along the temporal axis:
\[
F(V)=\{f_1,\dots,f_N\}.
\]
In practice, $N$ is a preset hyperparameter and the sampling indices are evenly spaced over the video duration.

\paragraph{Step 2: CLIP image embeddings (offline cache).}
For each sampled frame $f_j$, we compute a normalized CLIP image embedding
\begin{equation}
v_j = \frac{f_{\text{img}}(f_j)}{\|f_{\text{img}}(f_j)\|_2}\in\mathbb{R}^{d},\quad j=1,\dots,N,
\end{equation}
where $f_{\text{img}}(\cdot)$ is the CLIP image encoder.
The frame embeddings $\{v_j\}_{j=1}^{N}$ are precomputed and stored on disk to avoid recomputation during RL training.

\paragraph{Step 3: Sentence splitting of $y^{v}$.}
We split $y^{v}$ into $M$ sentences $\{s_m\}_{m=1}^{M}$ using a deterministic rule (e.g., newline or punctuation-based segmentation).
We drop empty segments after trimming whitespace.
If no valid sentence remains (i.e., $M=0$), we set $r_{\text{vis}}^{\text{TopK}}(V,y^{v})=0$.

\paragraph{Step 4: CLIP text embeddings (per sentence).}
For each sentence $s_m$, we compute a normalized CLIP text embedding
\begin{equation}
u_m=\frac{f_{\text{text}}(s_m)}{\|f_{\text{text}}(s_m)\|_2}\in\mathbb{R}^{d},\quad m=1,\dots,M,
\end{equation}
where $f_{\text{text}}(\cdot)$ is the CLIP text encoder.

\paragraph{Step 5: Similarity matrix and Top-$K$ aggregation.}
We form the frame--sentence similarity matrix $S\in\mathbb{R}^{N\times M}$:
\begin{equation}
S_{j,m}=v_j^\top u_m,
\end{equation}
which equals cosine similarity due to the $\ell_2$ normalization.
For each sentence $s_m$, we take the average over the Top-$K$ frame similarities:
\begin{equation}
\begin{aligned}
r(s_m)&=\frac{1}{K}\sum_{j\in\mathrm{TopK}(S_{:,m})} S_{j,m},\\
K&=\max\bigl(1,\lfloor \alpha N\rfloor\bigr),
\end{aligned}
\end{equation}
where $\alpha\in(0,1]$ controls the Top-$K$ ratio and $\mathrm{TopK}(S_{:,m})$ returns the indices of the $K$ largest entries in the $m$-th column. In our implementation, we set the Top-$K$ ratio to a fixed constant $\alpha=0.2$. i.e., we average over the most relevant $20\%$ frames per sentence, which empirically provides a robust trade-off between suppressing noisy/irrelevant frames while still retaining sufficient temporal evidence when the visual support is distributed across multiple moments.
Finally, we average over sentences to obtain
\begin{equation}
r_{\text{vis}}^{\text{TopK}}(V,y^{v})=\frac{1}{M}\sum_{m=1}^{M} r(s_m).
\end{equation}

\paragraph{Remarks.}
The Top-$K$ aggregation is designed for cases where evidence is distributed across multiple frames and multiple semantic fragments within $y^{v}$, and it reduces sensitivity to single-frame noise compared to CLIP-Max.

\section{Implementation Details of Lexical PCG Sanitization}
\label{pcg lexical safeguard}

This appendix provides the detailed computation of the lexical cleaning function $\mathrm{Clean}_{\mathrm{lex}}(y^t,y^a)$ and the copy penalty $L_{\mathrm{copy}}(y^t,y^a)$ used in the PCG reward.

\paragraph{Motivation.}
The PCG reward evaluates whether the intermediate perception--reasoning block increases the likelihood of the ground-truth next utterance. However, if the generated \texttt{<think>} block contains answer-like surface forms, the likelihood gain may be artificially increased by answer-cue tokens rather than genuine perception--cognition. Therefore, before computing PCG, we remove reasoning sentences that lexically copy the model's own generated \texttt{<answer>} block and subtract a copy penalty.

\paragraph{Lexical normalization.}
Given a sampled completion $y$, we extract the reasoning segment $y^t=\mathrm{Ext}_{t}(y)$ and the generated answer segment $y^a=\mathrm{Ext}_{a}(y)$. Before overlap computation, both segments are normalized by lowercasing, removing XML-like tags, normalizing whitespace, and removing punctuation. We do not apply semantic similarity or paraphrase matching, because useful reasoning should naturally be semantically related to the final answer; our goal is only to prevent lexical answer copying.

\paragraph{Sentence splitting.}
We split the reasoning segment into $M$ sentences:
\begin{equation}
y^t=\{s_1,\dots,s_M\}.
\end{equation}
If no valid reasoning sentence is extracted, we set $\mathrm{Clean}_{\mathrm{lex}}(y^t,y^a)=\emptyset$ and $L_{\mathrm{copy}}(y^t,y^a)=1$.

\paragraph{Exact $n$-gram overlap.}
Let $G_n(s_i)$ and $G_n(y^a)$ denote the token-level exact $n$-gram sets of the reasoning sentence $s_i$ and the generated answer $y^a$, respectively. We define
\begin{equation}
I_n(s_i,y^a)
=
\mathbb{I}\left[
G_n(s_i)\cap G_n(y^a)\neq \emptyset
\right].
\end{equation}
In practice, we use exact 4-gram overlap as a high-precision signal of answer copying and exact 3-gram overlap as a softer signal combined with ROUGE-L.

\paragraph{Answer-recall ROUGE-L.}
To detect cases where a reasoning sentence covers most of the generated answer in order, we define answer-recall ROUGE-L:
\begin{equation}
R_L(s_i,y^a)
=
\frac{\mathrm{LCS}(s_i,y^a)}
{\max(1,|y^a|)},
\end{equation}
where $\mathrm{LCS}(\cdot,\cdot)$ denotes the longest common subsequence length after lexical normalization, and $|y^a|$ is the number of tokens in the generated answer.

\paragraph{Answer-cue detection.}
A reasoning sentence is marked as containing answer-cue tokens if
\begin{equation}
\begin{aligned}
\ell_i =
\mathbb{I}\big[&
I_4(s_i,y^a)=1 \\
&\vee
\big(I_3(s_i,y^a)=1
\wedge R_L(s_i,y^a)>\tau_L\big) \\
&\vee
R_L(s_i,y^a)>\tau_H
\big].
\end{aligned}
\end{equation}
where we set $\tau_L=0.5$ and $\tau_H=0.8$. The first condition captures near-verbatim phrase copying. The second condition avoids over-penalizing short common 3-grams unless the answer-recall overlap is also substantial. The third condition detects cases where the reasoning sentence covers most of the generated answer even if exact long $n$-gram overlap is weakened by minor wording changes.

\paragraph{Lexical cleaning.}
We remove the detected answer-cue sentences from the reasoning block:
\begin{equation}
\mathrm{Clean}_{\mathrm{lex}}(y^t,y^a)
=
\bigoplus_{i:\ell_i=0}s_i.
\end{equation}
The cleaned intermediate block used for PCG scoring is then
\begin{equation}
\bar{z}(y)
=
y^v
\oplus
\mathrm{Clean}_{\mathrm{lex}}(y^t,y^a).
\end{equation}

\paragraph{Copy penalty.}
The copy penalty is the fraction of reasoning sentences that contain answer-cue tokens:
\begin{equation}
L_{\mathrm{copy}}(y^t,y^a)
=
\begin{cases}
\frac{1}{M}\sum_{i=1}^{M}\ell_i, & M>0,\\
1, & M=0.
\end{cases}
\end{equation}
This penalty does not introduce an additional weighting hyperparameter. It is zero when no lexical copying is detected and increases with the proportion of answer-cue sentences in the reasoning block.

\section{Details of BERTScore Semantic Reward}
\label{bertscore detail}

This appendix provides the detailed computation of the open-ended semantic reward
$r_{\text{sem}}(y^{a},a^{*})$ used in \cref{bertscore sec}.

\paragraph{Extraction and empty handling.}
Let $\hat{a}=y^{a}=\mathrm{Ext}_{a}(y)$ be the extracted \texttt{<answer>} segment of a sampled completion $y$, and let
$a^{*}=\mathrm{Ext}_{a}(y^{*})$ be the extracted \texttt{<answer>} segment of the reference structured output $y^{*}$.
If $y^{a}=\emptyset$ or $a^{*}=\emptyset$, we set $r_{\text{sem}}(y^{a},a^{*})=0$ by the empty-segment convention in \cref{method overview}.

\paragraph{Tokenization and contextual embeddings.}
We tokenize the candidate answer $y^{a}$ into $\{\hat{t}_i\}_{i=1}^{|\hat{a}|}$ and the reference answer $a^{*}$ into $\{t^{*}_j\}_{j=1}^{|a^{*}|}$.
A pretrained Transformer encoder $E(\cdot)$ (DeBERTa-XLarge-MNLI in our implementation) maps tokens to contextual embeddings:
\[
\small
\begin{aligned}
h^{\hat{a}}_i &= E(\hat{t}_i;\hat{a})\in\mathbb{R}^{D},\\
h^{a^{*}}_j &= E(t^{*}_j;a^{*})\in\mathbb{R}^{D}.
\end{aligned}
\]
We use cosine similarity between token embeddings:
\[
\mathrm{cos}(h,h')=\frac{h^\top h'}{\|h\|_2\|h'\|_2}.
\]

\paragraph{BERTScore precision, recall, and $F_1$.}
Following the standard BERTScore definition, we compute
\begin{align}
\small
P(\hat{a},a^{*})
&=\frac{1}{|\hat{a}|}
  \sum_{i=1}^{|\hat{a}|}
  \max_{j}\ \mathrm{cos}\!\left(h^{\hat{a}}_i,h^{a^{*}}_j\right),\label{eq:bertscore_p}\\
R(\hat{a},a^{*})
&=\frac{1}{|a^{*}|}
  \sum_{j=1}^{|a^{*}|}
  \max_{i}\ \mathrm{cos}\!\left(h^{\hat{a}}_i,h^{a^{*}}_j\right).\label{eq:bertscore_r}
\end{align}
and the harmonic mean
\begin{equation}
\mathrm{BERTScore\text{-}F1}(\hat{a},a^{*})
=
\frac{2P(\hat{a},a^{*})R(\hat{a},a^{*})}{P(\hat{a},a^{*})+R(\hat{a},a^{*})}.
\end{equation}

\paragraph{Reward value.}
The semantic reward is the clipped BERTScore-$F_1$:
\begin{equation}
r_{\text{sem}}(y^{a},a^{*})
=
\mathrm{clip}_{[0,1]}\Bigl(\mathrm{BERTScore\text{-}F1}(y^{a},a^{*})\Bigr).
\end{equation}

\paragraph{Implementation specifics.}
In our implementation we use the \texttt{bert-score} library with
\texttt{idf=False} and \texttt{rescale\_with\_baseline=False}, and we serve the scorer as a lightweight local service to avoid repeated model loading overhead during RL training.

\section{Computation Details of Dense Format Reward}
\label{dense reward}
This appendix specifies the dense format reward $r_{\text{fmt}}(y)$ used in \cref{acc reward}.

\paragraph{Tag set.}
Let $\mathcal{T}=\{\texttt{<perception>},\allowbreak \texttt{</perception>},\allowbreak \texttt{<think>},\allowbreak \texttt{</think>},\allowbreak \texttt{<answer>},\allowbreak \texttt{</answer>}\}$ be the required tag set.
Let $\mathrm{count}(y,\tau)$ denote the number of occurrences of tag $\tau$ in the raw completion string $y$.

\paragraph{(i) Tag existence score.}
We assign $+0.5$ if a required tag appears exactly once, and $-0.5$ otherwise:
\begin{equation}
S_{\text{tag}}(y)=\sum_{\tau\in\mathcal{T}}
\begin{cases}
+0.5,& \mathrm{count}(y,\tau)=1,\\
-0.5,& \text{otherwise.}
\end{cases}
\end{equation}

\paragraph{(ii) Structural order score.}
We add $+1.0$ if the tag order matches the required stage order
\texttt{<perception>...\ </perception>}\,$\rightarrow$\,\allowbreak
\texttt{<think>...\ </think>}\,$\rightarrow$\,\allowbreak
\texttt{<answer>...\ </answer>}
(without introducing a new tag between stage blocks), and add $0$ otherwise:
\begin{equation}
S_{\text{ord}}(y) = 
\begin{cases} 
+1.0, & \text{if } y \text{ follows strict stage order,} \\
0, & \text{otherwise.}
\end{cases}
\end{equation}

\paragraph{(iii) Boundary score.}
We enforce that $y$ starts with \texttt{<perception>} and ends with \texttt{</answer>} (allowing for leading/trailing whitespace).
If a boundary condition is met, we add $+0.5$; otherwise we add $-1.0$:
\begin{equation}
\small
\begin{aligned}
S_{\text{bd}}(y) ={}&
\begin{cases}
+0.5, & y \text{ starts with } \texttt{<perception>}, \\
-1.0, & \text{otherwise}
\end{cases} \\
&+ 
\begin{cases}
+0.5, & y \text{ ends with } \texttt{</answer>}, \\
-1.0, & \text{otherwise}.
\end{cases}
\end{aligned}
\end{equation}

\paragraph{Total format reward.}
The dense format reward is the sum of the three components:
\begin{equation}
r_{\text{fmt}}(y)=S_{\text{tag}}(y)+S_{\text{ord}}(y)+S_{\text{bd}}(y).
\end{equation}
This dense design yields informative scores that distinguish partially-correct formatting from fully invalid outputs, providing smoother optimization signals than a binary constraint.

\section{GRPO Objective}
\label{grpo}
We optimize the policy $\pi_{\theta}$ using Group Relative Policy Optimization (GRPO).
For each input prompt $x$, we draw a group of $G$ completions
$\{y_g\}_{g=1}^{G} \sim \pi_{\theta_{\text{old}}}(\cdot \mid x)$, where $\pi_{\theta_{\text{old}}}$ denotes the behavior (old) policy.
Let $y_g=(y_{g,1},\dots,y_{g,T_g})$ be the token sequence of the $g$-th completion with length $T_g$.
We define the token-wise importance ratio
\begin{equation}
r_{g,t}(\theta)
=
\frac{\pi_{\theta}(y_{g,t}\mid x, y_{g,<t})}{\pi_{\theta_{\text{old}}}(y_{g,t}\mid x, y_{g,<t})}.
\end{equation}

The scalar advantage $A_g$ is computed from the multi-dimensional reward vector
$\mathbf{r}_g=[r_{\text{sem}}(y_g^{a},a^*), \allowbreak r_{\text{fmt}}(y_g), \allowbreak r_{\text{vis}}(V,y_g^{v}), \allowbreak r_{\text{pcg}}(x,y_g^{v}\oplus y_g^{t})]$ via per-dimension group normalization and a weighted sum (see \cref{combine rewards}).

GRPO maximizes the following clipped surrogate objective with a KL regularizer:
\begin{equation}
\small
\begin{aligned}
\mathcal{J}_{\text{GRPO}}(\theta)
={}&
\mathbb{E}_{x\sim\mathcal{D}}
\Bigg[
\frac{1}{G}\sum_{g=1}^{G}\frac{1}{T_g}\sum_{t=1}^{T_g} \\
&\min\!\Big(
r_{g,t}(\theta)A_g,\\
&\qquad \mathrm{clip}\!\big(r_{g,t}(\theta),1-\epsilon,1+\epsilon\big)A_g
\Big) \\
&-\beta\,\mathrm{KL}\!\big(\pi_{\theta}(\cdot\mid x)\,\|\,\pi_{\text{ref}}(\cdot\mid x)\big)
\Bigg],
\end{aligned}
\end{equation}
where $\epsilon$ is the PPO-style clipping parameter and $\beta>0$ controls the KL penalty coefficient.
Following the standard GRPO formulation \cite{Shao2024DeepSeekMathPT}, we approximate the KL divergence using the estimator proposed by \citet{Schulman2017ProximalPO}:
\begin{equation}
\small
\begin{aligned}
\widehat{\mathrm{KL}}_{g,t}
={}&
\frac{\pi_{\text{ref}}(y_{g,t}\mid x,y_{g,<t})}{\pi_{\theta}(y_{g,t}\mid x,y_{g,<t})}\\
&-\log\frac{\pi_{\text{ref}}(y_{g,t}\mid x,y_{g,<t})}{\pi_{\theta}(y_{g,t}\mid x,y_{g,<t})}
-1.
\end{aligned}
\end{equation}
The final loss is computed by averaging this term over the group and token positions, i.e., $\mathcal{L}_{\text{GRPO}}(\theta)=-\mathcal{J}_{\text{GRPO}}(\theta)$.

\section{Cold-Start CoT Construction Prompt}
\label{sft cot}
\begin{fancybox}{System Prompt: Generate CoT (See-Think) Logic Analysis}
\footnotesize\raggedright
\setlength{\parskip}{0.5em}

You are an expert \textbf{Role-Play Logic Analyst}.
\textbf{YOUR GOAL}: You are NOT here to chat. Your goal is to \textbf{reverse-engineer the internal mental process} (Vision \& Thinking) of a specific character (the ASSISTANT) that logically leads to a provided \textbf{TARGET RESPONSE}.

\textbf{\#\#\# 1. GUIDELINES}
\begin{itemize}[leftmargin=*, topsep=0pt, itemsep=2pt]
    \item \textbf{Be Concise}: Focus only on critical reasoning. Avoid polite fillers.
    \item \textbf{NO Redundancy}: Briefly reference visual clues in the thinking step, do not re-describe them fully.
\end{itemize}

\textbf{\#\#\# 2. THE SHARED REALITY (Crucial Rule)}
The video represents a \textbf{Live Event} that occurred immediately before (or during) the current conversation.
\begin{itemize}[leftmargin=*, topsep=0pt, itemsep=2pt]
    \item \textbf{Physical Presence}: Both User and Assistant are physically present in this scene, fully involved in the context.
    \item \textbf{The Identity Separation Rule (NO FORCED BINDING)}: 
    \begin{itemize}
        \item The figures visible in the video \textbf{might NOT be} the User or Assistant, they could be the other people present. The User/Assistant might be standing just off-camera.
        \item \textbf{But At least ONE} of the speakers (User or Assistant) was definitely \textbf{ON SCREEN} experiencing the event directly.
        \item The other speaker was either also on screen, OR standing right next to the action witnessing it.
        \item \textbf{Strict Constraint}: Do NOT forcefully assume the visible figures are the speakers. 
        \item \emph{However}: The User and Assistant are \textbf{NOT random observers}. They are deeply connected to this event (either experiencing it directly or witnessing it from right next to the action).
    \end{itemize}
\end{itemize}

\textbf{\#\#\# 3. CORE RESPONSE LOGIC (The Flow)}
You must determine the Assistant's response based on the \textbf{Dialogue Direction} established in the history:
\begin{itemize}[leftmargin=*, topsep=0pt, itemsep=2pt]
    \item \textbf{Direction A (Topic Continuation)}: The dialogue directly \textbf{CONTINUES the conversation or interaction} shown in the video. (e.g., The video shows people talking/interacting; the current dialogue picks up right where the video left off, continuing the same specific topic).
    \item \textbf{Direction B (Inquiry/Reflection)}: The dialogue is a \textbf{Reaction/Inquiry} regarding the video event. (e.g., One person is asking the other about their thoughts, feelings, or reasons behind what they just did/said in the video).
\end{itemize}

\textbf{\#\#\# 4. THE PERSONALITY ANALYSIS}
Do NOT just output generic emotions. You must filter the observable reality through the Assistant's Profile:
\begin{itemize}[leftmargin=*, topsep=0pt, itemsep=2pt]
    \item \textbf{Standard Rule}: In normal situations, apply standard traits (e.g., A humorous character makes jokes).
    \item \textbf{High-Stakes Rule (Crucial)}: Adjust for the \textbf{intensity} of the atmosphere.
    \begin{itemize}
        \item If a humorous character faces \textbf{mortal danger}, they don't tell stand-up jokes $\to$ their humor becomes nervous, OR they drop the jokes to show unexpected \textbf{bravery}.
        \item If a wise character faces \textbf{tragedy}, they don't lecture $\to$ their wisdom becomes gentle silence.
    \end{itemize}
\end{itemize}

\textbf{\#\#\# OUTPUT FORMAT (Strict Step-by-Step)}
You must output XML-like tags in this exact order:

\textbf{\textless{}perception\textgreater}\\
Describe the \textbf{EVENT} objectively. \textbf{STRICTLY PROHIBITED: naming figures.}
\begin{enumerate}[leftmargin=*, topsep=0pt, itemsep=2pt]
    \item \textbf{The Core Event (Action \& Expression)}: Describe specific interactions (e.g., `People are having a tense discussion'). \textbf{Expression Check}: Describe visible emotions.
    \item \textbf{Key Objects}: Identify items driving the plot.
    \item \textbf{Atmosphere}: Describe the tension level (Safe vs. Dangerous) and lighting/vibe strictly to set the scene's emotional baseline.
\end{enumerate}
\textbf{\textless{}/perception\textgreater}

\textbf{\textless{}think\textgreater}\\
Synthesize Vision + Dialogue History to determine the response:
\begin{enumerate}[leftmargin=*, topsep=0pt, itemsep=2pt]
    \item \textbf{Analyze Vision}: What is the physical reality?
    \item \textbf{Analyze Dialogue History}: Look at the CONTEXT (Continuing topic vs. Discussing aftermath).
    \item \textbf{Determine Topic}: Combine [Event] + [Dialogue Direction].
    \item \textbf{Drafting (Personality Filter)}: Internalize ASSISTANT-\{assistant\_name\}'s mindset. Apply the \textbf{THE PERSONALITY ANALYSIS}. Check: If Vision is dangerous, does character show bravery/nervousness instead of casual traits?
\end{enumerate}
\textbf{\textless{}/think\textgreater}
\vspace{0.5em}

\textbf{CRITICAL}: Please stop generating immediately after \texttt{\textless{}/think\textgreater}. Do NOT generate the \texttt{\textless{}answer\textgreater} part, as that part already exists.
\end{fancybox}

\vspace{1em} 

\begin{fancybox}{User Prompt for Every Sample: Generate CoT (Look-Think) Logic Analysis}
\footnotesize\raggedright
\setlength{\parskip}{0.5em}

\textbf{Role}: You are analyzing the behavior of \textbf{\{assistant\_name\}} from `\{movie\}'.\\
\textbf{Context}: The character is physically present in the scene shown in the video.

\textbf{\#\#\# CHARACTER PROFILES}
USER (\{user\_name\}): \{user\_block\}

ASSISTANT (\{assistant\_name\}): \{assistant\_block\}

\textbf{\#\#\# INPUT CONTEXT}\\
\textbf{[Visual Reality]}: The raw recording of the immediate event.\\
\textbf{[Dialogue Context]}:\\
\{dialogue\_block\}

\textbf{\#\#\# TARGET RESPONSE (Ground Truth)}\\
The character accurately responded:\\
``\{target\_utterance\}''

\textbf{\#\#\# YOUR TASK}\\
Generate TWO blocks in order: \texttt{\textless{}perception\textgreater{}...\textless{}/perception\textgreater{}}, \texttt{\textless{}think\textgreater{}...\textless{}/think\textgreater{}}.

(1) \textbf{Step 1 \textless{}perception\textgreater{}...\textless{}/perception\textgreater{} (Camera Mode)}: Strictly describe \textbf{observable facts} ONLY inside these tags. Keep it concise.
\begin{itemize}[leftmargin=*, topsep=0pt, itemsep=2pt]
    \item 1. \textbf{The Core Event}: Describe objectively (no guessing). Do NOT name figures.
    \item 2. \textbf{Key Objects}: List plot-driving objects.
    \item 3. \textbf{Atmosphere}: State tension level.
    \item \textbf{NO ANALYSIS OR GUESSING HERE.} Just report what is seen.
\end{itemize}

(2) \textbf{Step 2 \textless{}think\textgreater{}...\textless{}/think\textgreater{} (Analytic Mode)}: Analyze visual clues and conversation context to explain the logic behind the target response.
\begin{itemize}[leftmargin=*, topsep=0pt, itemsep=2pt]
    \item 1. \textbf{Analyze Vision}: Summarize physical reality.
    \item 2. \textbf{Analyze Dialogue History}: Determine (A) Continuing topic or (B) Discussing reactions.
    \item 3. \textbf{Determine Topic}: Combine event + dialogue direction.
    \item 4. \textbf{Drafting (Personality Filter)}: Internalize \{assistant\_name\}'s mindset. Apply the \textbf{High-Stakes Personality Rule} (e.g., humor turns to bravery in danger) while thinking as the assistant.
\end{itemize}

\textbf{CRITICAL}:
\begin{itemize}[leftmargin=*, topsep=0pt, itemsep=2pt]
    \item \textbf{ANTI-COPY RULE (MANDATORY)}: In the \texttt{\textless{}think\textgreater} block, you MUST NOT copy, quote, or reproduce ANY exact wording from the TARGET RESPONSE / Ground Truth, including short phrases.
    \item Do NOT use quotation marks containing any target words.
    \item Do NOT repeat any contiguous 2+ words that appear in the TARGET RESPONSE.
    \item Instead, refer to the target response only by meaning, using abstract and probabilistic paraphrases (e.g., ``he would likely invite the other to lead first'', ``he signals deference'').
    \item After \texttt{\textless{}perception\textgreater}. \textbf{You MUST proceed to generate the \texttt{\textless{}think\textgreater} block}.
    \item Immediately after closing \texttt{\textless{}/think\textgreater}, please \textbf{stop generating}. Do NOT output the \texttt{\textless{}answer\textgreater} section.
\end{itemize}

\textbf{Start immediately with \textless{}vision\textgreater{}...}
\end{fancybox}

\section{Evaluation Metrics}
\label{metrics}
\begin{fancybox}{Prompt: Evaluation Metric - Visual Evidence Grounding (VEG)}
\footnotesize\raggedright
\setlength{\parskip}{0.5em}

Attention:
You are a ``Visual Evidence Auditor''. Your task is to evaluate the physical and factual alignment between the video and the dialogue.
This is a test of ``Observation Logic,'' not personality.

\{profile\_note\}

\textbf{Profile Context (for baseline temperament only):}\\
\{context\}

\textbf{Conversation History (for current emotional context):}\\
\{conversation\}

\textbf{Real Answer (Ground Truth): Reference only}:\\
\{gt\_answer\}

\textbf{Candidate Responses (ignore tags; judge spoken dialogue only):}\\
\{model\_responses\_block\}

Scoring (0-100), compute 3 subscores then weighted total:

\textbf{EVALUATION CRITERIA}:
\begin{enumerate}[leftmargin=*, topsep=0pt, itemsep=2pt]
    \item \textbf{Visual Triggering (50\%)}: Does the dialogue contain a direct, logical link to a specific action, object, or expression shown in the video?
    \item \textbf{Object/Scene Integrity (30\%)}: Does the response respect the physical limits of the scene? Penalize models mentioning invisible items.
    \item \textbf{Temporal Realism (20\%)}: Does the utterance match the immediacy of visual perception? Real-time visual grounding is typically reactive and sharp. Penalize excessive length if it drifts into ``describing the video'' instead of ``reacting to the video''.
\end{enumerate}

\textbf{SCORING ANCHORS (5 Tiers)}:
\begin{itemize}[leftmargin=*, topsep=0pt, itemsep=2pt]
    \item 0-20 (Tier 1): Hallucinates elements not in video or provides AI refusal.
    \item 21-40 (Tier 2): Overly descriptive/bookish; describes the scene rather than being in it.
    \item 41-60 (Tier 3): Factually safe but ``video-blind''; relies purely on text history.
    \item 61-80 (Tier 4): Shows clear awareness of the visual environment; situated and reactive.
    \item 81-100 (Tier 5): Precision sensing; captures micro-interactions only possible via vision.
\end{itemize}

\textbf{Output Format}:\\
For EACH model:
[[Model Name]]\\
Subscores: Triggering=\textless{}0-100\textgreater{}, Integrity=\textless{}0-100\textgreater{}, Temporal=\textless{}0-100\textgreater{}\\
Score: \textless{}0-100\textgreater{}\\
Reason: \textless{}explain how the response reflects visual precision vs. descriptive hallucination\textgreater{}\\
\{force\_complete\_msg\}
\end{fancybox}

\vspace{1em}

\begin{fancybox}{Prompt: Evaluation Metric - Situational Persona Compatibility (SPC)}
\footnotesize\raggedright
\setlength{\parskip}{0.5em}

Attention:
You are an expert Casting Director and Acting Coach.
Your task is to evaluate \{num\_models\} candidate responses.
Your goal is to evaluate if the line sounds like the character would ACTUALLY speak in the specific atmosphere shown in the video.

\textbf{EVALUATION PHILOSOPHY}:
\begin{itemize}[leftmargin=*, topsep=0pt, itemsep=2pt]
    \item \textbf{Dynamic Persona (Core Insight)}: Character consistency is NOT static. A ``humorous'' character in mortal danger should show nervous grit or brevity, NOT tell jokes. A ``wise'' leader in a high-pressure scene should be decisive and sharp, NOT deliver a poetic lecture.
    \item \textbf{The ``Presence'' Test}: Does the model sound like someone physically in the room?
    \item \textbf{Authenticity over Caricature}: Avoid rewarding models that simply use catchphrases or ``lore-dumping'' if it breaks the immediate scene tension.
\end{itemize}

\{profile\_note\}

\textbf{Character Profile \& Values}:\\
\{context\}

\textbf{Conversation History}:\\
\{conversation\}

\textbf{Real Answer (Ground Truth): Just for reference only}:\\
\{gt\_answer\}

\textbf{Candidate Responses (may include extra tags; apply the rule above)}:\\
\{model\_responses\_block\}

Scoring (0-100), compute 3 subscores and then a weighted total:

\begin{enumerate}[leftmargin=*, topsep=0pt, itemsep=2pt]
    \item \textbf{Situational Value Expression (40\%)}:
    \begin{itemize}
        \item Does the character express their core identity THROUGH the lens of the current event?
        \item \textbf{Crucial}: High scores go to characters who adapt their expression. (e.g., A proud villain showing rare caution when the video shows an overwhelming threat).
        \item Penalize ``Static Labeling'' where characters repeat slogans regardless of the visual pressure.
    \end{itemize}
    \item \textbf{Acting Realism \& Rhythm (30\%)}:
    \begin{itemize}
        \item \textbf{Hard Penalty}: Over-dramatized monologues or ``stagey'' declamations that destroy the scene's tension.
        \item \textbf{Focus}: Does the response respect the \textbf{pacing of the moment}? High-stakes/urgent scenes usually require sharp, reactive lines, long-winded explanations in tense moments break immersion.
    \end{itemize}
    \item \textbf{Sensory-Driven Modulation (30\%)}:
    \begin{itemize}
        \item Does the voice feel like it was filtered through the ``Eyes''?
        \item If the video shows the character is exhausted, is the line shorter and lower energy?
    \end{itemize}
\end{enumerate}

\textbf{Scoring Anchors (5 Tiers)}:
\begin{itemize}[leftmargin=*, topsep=0pt, itemsep=2pt]
    \item \textbf{0-20 (Tier 1: Disconnected)}: Meta-responses, refusals, or completely out-of-universe logic.
    \item \textbf{21-40 (Tier 2: Static Caricature)}: Sounds like a generic fan-fiction description. Overly wordy, ignores the video's urgency, or uses ``book-narrator'' prose instead of speech.
    \item \textbf{41-60 (Tier 3: Rigid Profile)}: Accurate to the text-only Profile, but ``blind'' to the video. The character acts the same in a forest battle as they would in a classroom.
    \item \textbf{61-80 (Tier 4: Believable Actor)}: Consistent and shows clear awareness of the scene. The tone shifts correctly with the visual tension. Minor style gaps.
    \item \textbf{81-100 (Tier 5: Immersive Soul)}: Elite performance. The line is concise, perfectly situated, and captures the character's unique voice under the \emph{exact} pressure shown in the video.
\end{itemize}

Output (for EACH model):
[[Model Name]]\\
Subscores: Values=\textless{}0-100\textgreater{}, Style=\textless{}0-100\textgreater{}, Stakes=\textless{}0-100\textgreater{}\\
Score: \textless{}0-100\textgreater{}\\
Reason: \textless{}Focus on how the character's static profile evolved/modulated in response to the specific video atmosphere\textgreater{}\\
\{force\_complete\_msg\}
\end{fancybox}

\vspace{1em}

\begin{fancybox}{Prompt: Evaluation Metric - Conversational Naturalism (CN)}
\footnotesize\raggedright
\setlength{\parskip}{0.5em}

Attention:
You are a ``Linguistic Stylist''. Your task is to evaluate the ``Spoken Quality'' of the response.
Is it a real person talking, or an AI writing a book?

\{profile\_note\}

\textbf{Character Profile (for linguistic habits):}\\
\{context\}

\textbf{Conversation History:}\\
\{conversation\}

\textbf{Real Answer (Ground Truth): Reference only}:\\
\{gt\_answer\}

\textbf{Candidate Responses (judge ONLY the spoken line):}\\
\{model\_responses\_block\}

Scoring (0-100), compute 2 subscores then weighted total:

\textbf{EVALUATION CRITERIA}:
\begin{enumerate}[leftmargin=*, topsep=0pt, itemsep=2pt]
    \item \textbf{Oral Realism (60\%)}: Use of colloquial phrasing and natural sentence structures.
    \item \textbf{Anti-Narrative Voice (40\%)}: Penalize stage directions (\emph{smiles}), ``he said'', or bookish monologues.
\end{enumerate}

\textbf{SCORING ANCHORS (5 Tiers)}:
\begin{itemize}[leftmargin=*, topsep=0pt, itemsep=2pt]
    \item 0-20 (Tier 1): Script/AI Voice; feels like a robot or contains ``As an AI...''.
    \item 21-40 (Tier 2): Narrative Prose; structured like a book paragraph, too formal/polished for speech.
    \item 41-60 (Tier 3): Formal Dialogue; grammatically correct but stiff, like a formal interview.
    \item 61-80 (Tier 4): Colloquial Spoken; good flow, uses contractions and natural slang.
    \item 81-100 (Tier 5): Situated Utterance; feels like a ``snippet'' of real life—short and impactful.
\end{itemize}

\textbf{Output Format}:\\
For EACH model:
[[Model Name]]\\
Subscores: Oral=\textless{}0-100\textgreater{}, Anti-Narrative=\textless{}0-100\textgreater{}\\
Score: \textless{}0-100\textgreater{}\\
Reason: \textless{}explain how the response avoids narrator-voice and achieves natural speech\textgreater{}\\
\{force\_complete\_msg\}
\end{fancybox}

\section{Visual-Augmented Text-only Role-Playing Baselines}
\label{app:visual_augmented_rp}

To provide a fairer comparison with text-only role-playing agents, we construct visual-augmented variants of RoleMRC, Crab, and Haruhi. During inference, we first use Gemini-3-Pro-Preview to convert each video into a textual visual description, and then provide the same description as an additional conditioning input to the prompt of each text-only role-playing model. We subsequently re-evaluate all systems using the same GPT-5-mini evaluation protocol. The complete results are reported in Table~\ref{tab:visual_augmented_results}.

\begin{table}[!t]
    \centering
    \small
    \setlength{\tabcolsep}{2.0pt}
    \renewcommand{\arraystretch}{1.02}
    \begin{tabular}{@{}lccc|c@{}}
        \toprule
        Model & VEG$\uparrow$ & SPC$\uparrow$ & CN$\uparrow$ & Avg.$\uparrow$ \\
        \midrule
        GPT-5                         & 79.87 & 79.64 & 81.47 & 80.33 \\
        Gemini-3                      & 80.62 & 77.41 & 83.03 & 80.35 \\
        \midrule
        InternVL3-38B-Instruct        & 74.98 & 70.52 & 75.32 & 73.61 \\
        Qwen2.5-VL-32B-Instruct       & 74.63 & 70.78 & 74.60 & 73.34 \\
        InternVL3-14B-Instruct        & 73.81 & 67.94 & 73.50 & 71.75 \\
        \midrule
        InternVL3-8B-Instruct         & 71.64 & 66.38 & 71.55 & 69.86 \\
        Qwen2.5-VL-7B-Instruct        & 70.69 & 65.74 & 72.13 & 69.52 \\
        RoleMRC                       & 71.46 & 66.83 & 73.40 & 70.56 \\
        \textbf{RoleMRC-visual}       & \textbf{72.27} & \textbf{65.33} & \textbf{72.46} & \textbf{70.02} \\
        Crab                          & 69.38 & 66.00 & 76.09 & 70.49 \\
        \textbf{Crab-visual}          & \textbf{71.05} & \textbf{65.83} & \textbf{76.71} & \textbf{71.20} \\
        Haruhi                        & 67.51 & 62.82 & 71.22 & 67.18 \\
        \textbf{Haruhi-visual}        & \textbf{71.18} & \textbf{65.10} & \textbf{73.89} & \textbf{70.06} \\
        Qwen2.5-VL-7B-SFT             & 70.13 & 66.48 & 72.61 & 69.74 \\
        \rowcolor{cyan!18}
        Char-EBM-CLIP-TopK            & 73.37 & 69.88 & 74.82 & 72.69 \\
        \rowcolor{cyan!18}
        Char-EBM-CLIP-Max             & 74.40 & 70.02 & 75.90 & 73.44 \\
        \bottomrule
    \end{tabular}
    \caption{\textbf{Comparison with visual-augmented text-only role-playing baselines.}
    For visual variants, Gemini-3-Pro-Preview first converts the video into a textual visual description, which is then provided as additional conditioning input to the corresponding role-playing model. All systems are evaluated under the same GPT-5-mini protocol.}
    \label{tab:visual_augmented_results}
\end{table}

On VEG, visual augmentation improves RoleMRC, Crab, and Haruhi by \textbf{+0.81}, \textbf{+1.67}, and \textbf{+3.67}, respectively. This indicates that text-only role-playing models can exploit externally provided visual descriptions and incorporate more explicit scene-related cues into their final responses.

However, improved VEG does not consistently translate into higher SPC. RoleMRC decreases from \textbf{66.83 to 65.33}, while Crab slightly decreases from \textbf{66.00 to 65.83}. One possible explanation is that these text-only role-playing agents have strong static persona priors, whereas the additional visual description requires them to dynamically adjust their default behavior according to danger, tension, or other scene states. This additional conditioning may compete with their existing persona-conditioned generation patterns, while extra scene details may also introduce context noise.

The effect on CN is comparatively small. RoleMRC and Crab change by only \textbf{-0.94} and \textbf{+0.62}, respectively, suggesting that conversational naturalness is still mainly determined by the models' existing language generation and role-playing capabilities.

Haruhi is a notable exception, with gains of \textbf{+3.67/+2.28/+2.67} on VEG/SPC/CN. We interpret this primarily as a weak-baseline compensation effect rather than evidence of stronger visual understanding or visual--persona integration. Haruhi has the lowest original performance, particularly on VEG and SPC; explicit visual descriptions provide event, interaction, atmosphere, and risk information that reduces scene ambiguity and narrows the plausible response space, leading to larger improvements. Nevertheless, Char-EBM-CLIP-Max still achieves the strongest overall performance among these comparable systems, indicating that direct multimodal modeling and reinforcement learning provide benefits beyond simply appending externally generated visual descriptions.

\section{Human Evaluation and Validation of LLM-as-a-Judge}
\label{app:human_llm_validation}

To more thoroughly validate both our main experimental conclusions and the reliability of GPT-5-mini as an automatic evaluator, we conduct complementary human analyses at both the system and individual-sample levels. Specifically, we examine: (i) human-only pairwise comparisons derived from independent human ratings; (ii) system-level agreement between human and GPT-5-mini scores; (iii) metric-specific differences in their score distributions; and (iv) sample-level agreement after mapping both evaluations onto the same five-tier semantic scale.

\subsection{Human Evaluation Protocol and Human-only Pairwise Comparison}
\label{app:human_pairwise}

\paragraph{Human evaluation protocol.}
We stratified 300 test samples according to the distribution of the test set and invited five annotators who were familiar with the target films and the task definition. The annotators independently evaluated the actual generated responses of four pre-selected representative systems: EBM, Qwen2.5-VL-7B-SFT, the strongest VLM under the corresponding automatic evaluation metric, and the strongest text-only role-playing baseline under the corresponding metric.

Each response was independently assessed along the same three dimensions used in automatic evaluation: Visual Evidence Grounding (VEG), Situational Persona Compatibility (SPC), and Conversational Naturalism (CN). Human evaluation followed a five-level ordinal semantic rubric (Tier 1--5) aligned with the automatic evaluation criteria. For system-level aggregation, the five tiers were mapped to scores of 20, 40, 60, 80, and 100, respectively. Thus, each metric contains 1,200 evaluated model responses (300 samples $\times$ 4 systems). Automatic evaluation results were used only to pre-select competitive representative baselines for human evaluation; all human judgments were independently assigned by the annotators based directly on the actual generated responses.

\paragraph{Human-only sample-wise comparison.}
Beyond comparing system-level average human scores, we further derive sample-wise pairwise preferences solely from the human annotations. Since the human rubric consists of five ordered quality levels, we use the median rating across the five annotators as the \emph{human consensus tier} for each model--sample pair. The median preserves the ordinal structure of the rubric, avoids ambiguity caused by tied modes, and provides a relatively conservative consensus criterion for pairwise comparison.

For each sample and metric, EBM is counted as a \emph{Win}, \emph{Loss}, or \emph{Tie} if its human consensus tier is respectively higher than, lower than, or equal to that of the corresponding baseline. The resulting human-only comparisons are reported in Table~\ref{tab:human_pairwise}.

\begin{table*}[!t]
    \centering
    \begin{small}
    \setlength{\tabcolsep}{5.0pt}
    \renewcommand{\arraystretch}{1.05}
    \begin{tabular}{@{}lccc@{}}
        \toprule
        Comparison
        & VEG W/L/T (\%)
        & SPC W/L/T (\%)
        & CN W/L/T (\%) \\
        \midrule

        EBM vs.\ Qwen-VL-SFT
        & 56 / 22 / 22
        & 64 / 15 / 21
        & 40 / 28 / 32 \\

        EBM vs.\ strongest VLM
        & 30 / 34 / 36
        & 29 / 33 / 38
        & 37 / 33 / 30 \\

        EBM vs.\ strongest text-only RP
        & \cellcolor{cyan!18}\textbf{70 / 11 / 19}
        & \cellcolor{cyan!18}\textbf{76 / 8 / 16}
        & 32 / 34 / 34 \\

        \bottomrule
    \end{tabular}
    \end{small}
    \caption{
    \textbf{Human-only sample-wise pairwise comparison.}
    Win/Loss/Tie is determined solely from the median consensus tier of five human annotators for each sample. ``Strongest VLM'' and ``strongest text-only RP'' denote the highest-scoring baseline within the corresponding category under the automatic evaluation protocol for each metric.
    }
    \label{tab:human_pairwise}
\end{table*}

The human-only pairwise results reproduce the overall pattern observed in the system-level evaluation. Relative to Qwen-VL-SFT, EBM achieves higher human win rates than loss rates on all three dimensions, with the clearest advantage on SPC (64\% Win vs.\ 15\% Loss). Against the strongest VLMs, EBM remains broadly comparable on VEG and SPC and shows a slight advantage on CN, indicating that EBM approaches substantially larger vision-language models despite its smaller scale.

The most pronounced difference appears against the strongest text-only role-playing baselines. EBM achieves human win rates of \textbf{70\% on VEG} and \textbf{76\% on SPC}, with only 11\% and 8\% losses, respectively. In contrast, CN is nearly balanced (32\% Win, 34\% Loss, and 34\% Tie). This pattern suggests that strong text-only role-playing models retain competitive linguistic naturalness, but their lack of direct visual access substantially limits visual grounding and situational persona adaptation.

\subsection{System-level Consistency between Human and GPT-5-mini Evaluation}
\label{app:human_llm_system}

We next compare the aggregated human scores with GPT-5-mini scores for the same representative systems. The results are shown in Table~\ref{tab:human_llm_corr}. For each metric, the four values are ordered as EBM, Qwen2.5-VL-7B-SFT, the strongest VLM under that metric, and the strongest text-only role-playing baseline under that metric.

\begin{table*}[!t]
    \centering
    \begin{small}
    \setlength{\tabcolsep}{3.2pt}
    \renewcommand{\arraystretch}{1.0}
    \resizebox{0.95\textwidth}{!}{%
    \begin{tabular}{@{}lccc@{}}
        \toprule
        Metric
        & Human Score
        & GPT-5-mini Score
        & Pearson Correlation \\
        \midrule

        VEG
        & 83.3 / 68.9 / 85.2 / 54
        & 74.25 / 70.82 / 74.95 / 70.51
        & \cellcolor{cyan!18}\textbf{92.78\%} \\

        SPC
        & 82.1 / 60.3 / 83.8 / 49.2
        & 70.37 / 66.46 / 71.01 / 66.87
        & \cellcolor{cyan!18}\textbf{93.95\%} \\

        CN
        & 85.3 / 79.1 / 82.03 / 88.3
        & 74.78 / 72.08 / 74.43 / 75.23
        & \cellcolor{cyan!18}\textbf{89.43\%} \\

        \bottomrule
    \end{tabular}%
    }
    \end{small}
    \caption{
    \textbf{System-level consistency between human and GPT-5-mini evaluation.}
    For each metric, the four scores correspond to EBM, Qwen2.5-VL-7B-SFT, the strongest VLM, and the strongest text-only role-playing baseline under that metric, respectively.
    }
    \label{tab:human_llm_corr}
\end{table*}

The Pearson correlations are consistently high across VEG, SPC, and CN, indicating that the relative system-level trends produced by GPT-5-mini broadly align with those observed in human evaluation. Importantly, we do not rely on these four-system correlations alone. The human-only sample-wise comparison in Table~\ref{tab:human_pairwise} and the substantially larger sample-level tier analysis in Section~\ref{app:sample_tier_agreement} provide complementary evidence regarding the reliability of the automatic evaluator.

\subsection{Analysis of Score-scale Differences between Human and GPT-5-mini}
\label{app:human_llm_difference}

Although human and GPT-5-mini evaluations exhibit similar system-level trends, their absolute score gaps are not identical. We therefore further inspect the metric-specific differences.

\paragraph{\textbf{VEG.}}
For VEG, the human scores of the strongest VLM and the strongest text-only RP baseline differ by 31.2 points, whereas GPT-5-mini separates them by only 4.44 points. We attribute this compression primarily to several characteristics of LLM-based judging.

First, an LLM judge can reliably determine whether an utterance is broadly \emph{compatible} with the current scene, but in borderline cases it is less sensitive than human annotators to the distinction between a response that merely fits the scene and one that genuinely depends on visual evidence available in the current video. Second, GPT-5-mini has strong linguistic and world-knowledge priors. Consequently, when an utterance is semantically plausible within the story world, character relationship, and general situation, it may still receive a relatively favorable VEG score even if it does not exploit specific visual actions, expressions, or objects.

For example, an utterance such as ``Keep your wand ready.'' is semantically plausible in a Hogwarts setting. A human annotator, however, may further ask whether the current frames actually contain a visual trigger that would justify immediately raising or preparing a wand. Finally, GPT-5-mini tends to assign many responses without severe scene violations to a relatively concentrated middle-to-high score range, further reducing the numerical separation between visually grounded models and text-only role-playing systems.

These results therefore do not suggest that GPT-5-mini ignores visual grounding; rather, its distinction between \emph{scene compatibility} and \emph{genuine video dependence} is less fine-grained than that of human annotators.

\paragraph{\textbf{SPC.}}
The SPC discrepancy follows a closely related pattern. In human evaluation, the strongest VLM and strongest text-only RP baseline differ by 31.2 points on VEG and by an even larger 34.6 points on SPC, whereas the corresponding GPT-5-mini gaps are only 4.44 and 4.14 points. This does not imply that the LLM judge fails to recognize character consistency. Instead, it suggests that GPT-5-mini is less fine-grained when distinguishing between being generally compatible with a character profile and reproducing how a \emph{specific character would react in a specific situation}.

Under our evaluation setting, GPT-5-mini primarily judges character behavior with access to the explicitly provided character profile. In contrast, annotators familiar with the corresponding films possess richer character knowledge, including interpersonal relationships, prior behavioral patterns, habitual emotional expression, values, and characteristic reactions under different levels of pressure. Thus, even when two responses are both broadly compatible with the given profile, human annotators can further distinguish which response more faithfully reflects how that particular character would behave in the current scene.

This finer-grained judgment can magnify the difference between visual models and text-only role-playing models. Models with visual input can directly exploit current actions, facial expressions, object states, and situational pressure, making them more likely to produce character-specific reactions tightly coupled with the immediate situation. Text-only RP models, although often strong at maintaining global persona consistency, are more likely to fall back on generic profile traits or story-world priors.

For example, several characters may all appear brave or anxious in a dangerous situation, but character fidelity is expressed through \emph{how} that state manifests---through provocation, restraint, hesitation, or prioritizing the protection of companions. GPT-5-mini can recognize many such distinctions, but at a coarser granularity than annotators familiar with the films, thereby compressing the SPC gap between strong visual models and conventional text-only RP systems.

\paragraph{\textbf{CN.}}
CN exhibits a different pattern: human scores are generally higher than those assigned by GPT-5-mini. Our qualitative inspection suggests that GPT-5-mini is more conservative toward responses that are formal, literary, rhetorical, or relatively long. When an utterance resembles polished prose, a speech, or novelistic dialogue, GPT-5-mini tends to penalize it under a general conversational-naturalness criterion.

Human annotators familiar with the characters, however, can interpret such stylistic choices in the context of character-specific speech patterns and are therefore more accepting of formal language when it is appropriate for the role. Characters such as Dumbledore, Aragorn, or Snape naturally employ relatively formal, restrained, or rhetorical language.

The main CN discrepancy therefore does not arise because human annotators neglect naturalness. Rather, humans are more willing to accept formal expression when it is consistent with a character's established speaking style, whereas GPT-5-mini places relatively greater emphasis on generic conversational immediacy, colloquiality, and conciseness.

\subsection{Sample-level Tier Agreement}
\label{app:sample_tier_agreement}

Finally, we directly examine the relationship between human and GPT-5-mini judgments at the individual model--sample level using the original 300-sample human evaluation set. For each model--sample instance, we again use the median of the five annotator ratings as the human consensus tier. We then map GPT-5-mini's continuous 0--100 score on the same samples to Tier 1--5 using the same semantic tier boundaries defined by the evaluation rubric.

We categorize the difference between the two judgments into three cases: \emph{Exact Tier}, where both assign the same tier; \emph{Adjacent-Tier Difference}, where their tiers differ by exactly one level; and \emph{Severe Difference}, where their tiers differ by two or more levels. Results across the 1,200 model--sample instances for each metric are shown in Table~\ref{tab:sample_tier_agreement}.

\begin{table}[!t]
    \centering
    \small
    \setlength{\tabcolsep}{3.0pt}
    \renewcommand{\arraystretch}{1.03}
    \resizebox{\columnwidth}{!}{%
    \begin{tabular}{@{}lccc@{}}
        \toprule
        Metric
        & Exact Tier
        & Adjacent Tier
        & $\geq$2-Tier Difference \\
        \midrule

        VEG
        & 58.5\%
        & 35.0\%
        & 6.5\% \\

        SPC
        & 54.0\%
        & 36.5\%
        & 9.5\% \\

        CN
        & 63.0\%
        & 34.0\%
        & 3.0\% \\

        \bottomrule
    \end{tabular}%
    }
    \caption{
    \textbf{Sample-level agreement between human consensus and GPT-5-mini judgments.}
    Exact Tier denotes identical five-level quality judgments; Adjacent Tier denotes a one-tier difference; and $\geq$2-Tier Difference denotes a disagreement of two or more quality levels.
    }
    \label{tab:sample_tier_agreement}
\end{table}

For VEG, SPC, and CN, respectively, \textbf{93.5\%, 90.5\%, and 97.0\%} of model--sample judgments are either assigned to exactly the same tier or differ by only one adjacent quality level. Severe disagreements spanning two or more tiers account for only 6.5\%, 9.5\%, and 3.0\%, respectively.

Together with the human-only pairwise results and system-level correlations, this sample-level analysis provides a more complete picture of evaluation reliability. GPT-5-mini does exhibit systematic differences in score calibration---most notably, compressed VEG/SPC gaps and more conservative CN scoring---but the overwhelming majority of individual judgments remain within one semantic quality tier of the human consensus.

We therefore use GPT-5-mini primarily as a scalable evaluator of relative model quality, while treating the complementary human evaluation as validation of both the overall ranking trends and the interpretation of metric-specific differences.

\section{Cross-Judge Validation with Gemini}
\label{app:gemini_judge_validation}

To reduce reliance on a single GPT-5-mini judge, we further use the independent heterogeneous multimodal model Gemini-3-Flash-Preview to re-evaluate the full set of evaluated models. We keep the candidate outputs, video and dialogue contexts, VEG/SPC/CN rubrics, and model-anonymization protocol identical to those used with GPT-5-mini. The complete results are reported in Table~\ref{tab:gemini_judge_results}.

\begin{table}[!t]
    \centering
    \small
    \setlength{\tabcolsep}{2.0pt}
    \renewcommand{\arraystretch}{1.02}
    \begin{tabular}{@{}lccc|c@{}}
        \toprule
        Model & VEG$\uparrow$ & SPC$\uparrow$ & CN$\uparrow$ & Avg.$\uparrow$ \\
        \midrule
        GPT-5                         & 63.36 & 61.44 & 92.48 & 72.43 \\
        Gemini-3                      & 64.86 & 67.40 & 87.60 & 73.29 \\
        \midrule
        InternVL3-38B-Instruct        & 51.72 & 52.51 & 79.74 & 61.32 \\
        Qwen2.5-VL-32B-Instruct       & 52.17 & 52.22 & 78.76 & 61.05 \\
        InternVL3-14B-Instruct        & 50.41 & 48.71 & 77.25 & 58.79 \\
        \midrule
        InternVL3-8B-Instruct         & 48.83 & 45.51 & 73.88 & 56.07 \\
        Qwen2.5-VL-7B-Instruct        & 47.60 & 43.61 & 73.53 & 54.91 \\
        RoleMRC                       & 49.56 & 46.42 & 75.74 & 57.24 \\
        Crab                          & 48.74 & 46.80 & 78.34 & 57.96 \\
        Haruhi                        & 43.95 & 40.83 & 59.67 & 48.15 \\
        Qwen2.5-VL-7B-SFT             & 47.78 & 44.28 & 73.96 & 55.34 \\
        \rowcolor{cyan!18}
        Char-EBM-CLIP-TopK            & 51.55 & 50.12 & 78.20 & 59.96 \\
        \rowcolor{cyan!18}
        Char-EBM-CLIP-Max             & 52.35 & 52.40 & 79.60 & 61.45 \\
        \bottomrule
    \end{tabular}
    \caption{\textbf{Evaluation results using Gemini-3-Flash-Preview as an independent judge.} Candidate outputs, video and dialogue contexts, evaluation rubrics, and anonymization protocols are identical to those used with GPT-5-mini.}
    \label{tab:gemini_judge_results}
\end{table}

We observe metric-specific calibration differences between Gemini-3-Flash-Preview and GPT-5-mini. Gemini generally adopts a more conservative numerical scale on VEG and SPC, while CN does not exhibit the same directional reduction, suggesting a metric-specific calibration difference rather than a uniform global severity shift. For VEG, Gemini places greater emphasis on whether a response can be directly grounded in verifiable visual evidence, such as concrete actions, objects, or facial expressions. Responses that are semantically compatible with the scene but lack an explicit visual trigger are therefore more likely to remain in the middle score range. This indicates a stricter distinction between \emph{scene compatibility} and \emph{verifiable visual dependence}.

A similar pattern appears on SPC. Gemini evaluates not only whether a response broadly matches the character profile, but also whether the character's behavior is concretely modulated by the visual pressure of the current situation. Responses that are overly abstract, excessively dramatic, or insufficiently aligned with the current tension, expressions, or actions may therefore receive more conservative scores. Thus, the lower SPC scale mainly reflects stricter thresholds for \emph{situational modulation} and \emph{acting realism}.

Despite these differences in absolute score calibration, the two judges produce highly consistent relative model rankings, as shown in Table~\ref{tab:inter_judge_agreement}.

\begin{table}[!t]
    \centering
    \small
    \setlength{\tabcolsep}{2.2pt}
    \renewcommand{\arraystretch}{1.05}
    \resizebox{\columnwidth}{!}{%
    \begin{tabular}{@{}lcccc@{}}
        \toprule
        Metric
        & Pearson ($r$)
        & Spearman ($\rho$)
        & Kendall ($\tau$)
        & Pairwise Order Agreement \\
        \midrule
        VEG  & 0.924 & 0.912 & 0.769 & 88.5\% \\
        SPC  & 0.965 & 0.973 & 0.897 & 94.9\% \\
        CN   & 0.775 & 0.912 & 0.744 & 87.2\% \\
        \midrule
        \textbf{Avg.}
        & \textbf{0.986}
        & \textbf{0.978}
        & \textbf{0.923}
        & \textbf{96.2\%} \\
        \bottomrule
    \end{tabular}%
    }
    \caption{\textbf{Inter-judge agreement between GPT-5-mini and Gemini-3-Flash-Preview.} Pairwise Order Agreement measures the proportion of model pairs for which the two judges produce the same relative ordering.}
    \label{tab:inter_judge_agreement}
\end{table}

Overall, although GPT-5-mini and Gemini-3-Flash-Preview exhibit metric-specific differences in absolute score calibration, they remain highly consistent in model ranking and pairwise relative ordering. These cross-judge results substantially reduce the dependence of our main conclusions on the preference or numerical scale of a single automatic evaluator.

\section{Cross-Domain Validation on HoloAssist}
\label{app:holoassist_cross_domain}

Existing non-movie datasets generally do not provide the complete combination required by our video-grounded role-playing setting: aligned video, role information, dialogue history, and a grounded target utterance. Motivated by the reviewers' suggestion to evaluate beyond cinematic scenes, we therefore adapt HoloAssist~\citep{Wang_2023_ICCV} into a small-scale, quality-controlled 500-sample cross-domain pilot. We intentionally prioritize sample filtering and alignment quality over large-scale automatic construction to preserve reliable correspondence among video, dialogue history, role information, and target responses.

HoloAssist differs substantially from our cinematic benchmark. It records dyadic real-world task collaboration in which a Task Performer manipulates objects, devices, and components from a first-person viewpoint, while a remote Instructor observes the egocentric video and provides spoken guidance. Its interaction structure is therefore closer to first-person gameplay-like or embodied task interaction than cinematic narrative: visual states evolve continuously through hand--object manipulation, object selection, spatial positioning, task progress, mistakes, and corrections.

We reformulate the task as:
\begin{center}
\small
\textit{first-person video + dialogue history + session-specific behavioral profiles}\\
$\rightarrow$ \textit{next Instructor utterance}.
\end{center}
The Task Performer is treated as the user and the Instructor as the target assistant. We use only actual \texttt{Conversation} annotations, excluding Narration and action annotations. Consecutive same-speaker utterances are merged within local interaction episodes, only Instructor turns are used as prediction targets, and each input video is strictly cropped before the target utterance begins to avoid target-period leakage.

For persona information, we construct session-level behavioral profiles. For each unique session, the full Conversation transcript is used to generate one Task Performer profile and one Instructor profile, shared by all samples from that session. Profiles contain only high-level and stable observable traits, including personality, catchphrases, speech style, relationship dynamics, strengths, weaknesses, background, and motivation. They are explicitly prohibited from encoding specific task answers, exact action steps, object locations, device operations, or sample-specific instructions.

After filtering and quality control, we obtain a fixed 500-sample evaluation subset. Results are shown in Table~\ref{tab:main_results_holoassist}.

\begin{table}[!t]
    \centering
    \small
    \setlength{\tabcolsep}{2.0pt}
    \renewcommand{\arraystretch}{1.02}
    \begin{tabular}{@{}lccc|c@{}}
        \toprule
        Model & VEG$\uparrow$ & SPC$\uparrow$ & CN$\uparrow$ & Avg.$\uparrow$ \\
        \midrule
        InternVL3-38B-Instruct         & 67.65 & 62.56 & 81.44 & 70.55 \\
        Qwen2.5-VL-32B-Instruct        & 68.51 & 64.69 & 81.76 & 71.65 \\
        InternVL3-14B-Instruct         & 64.57 & 56.04 & 78.04 & 66.22 \\
        \midrule
        InternVL3-8B-Instruct          & 58.18 & 53.47 & 75.21 & 62.29 \\
        Qwen2.5-VL-7B-Instruct         & 61.40 & 55.16 & 76.61 & 64.39 \\
        RoleMRC                        & 58.86 & 56.53 & 78.91 & 64.77 \\
        Crab                           & 51.62 & 46.51 & 82.48 & 60.20 \\
        Haruhi                         & 53.19 & 45.72 & 69.33 & 56.08 \\
        \rowcolor{cyan!18}
        Char-EBM-CLIP-Max              & 66.06 & 57.30 & 82.04 & 68.47 \\
        \bottomrule
    \end{tabular}
    \caption{Cross-domain evaluation on the quality-controlled 500-sample HoloAssist pilot subset.}
    \label{tab:main_results_holoassist}
    \vskip -0.1in
\end{table}

Despite the substantial domain shift, Char-EBM-CLIP-Max remains highly competitive. Relative to its 7B base model Qwen2.5-VL-7B-Instruct, EBM improves VEG/SPC/CN/Avg. by +4.66/+2.14/+5.43/+4.08. Compared with the larger InternVL3-14B, it improves VEG/SPC/CN/Avg. by +1.49/+1.26/+4.00/+2.25. Against the strongest text-only role-playing baseline RoleMRC, EBM improves VEG by \textbf{+7.20} and the average score by \textbf{+3.70}.

The domain shift spans multiple dimensions: third-person cinematic shots become egocentric first-person video; actors and movie scenes become hands, tools, devices, spatial relations, and evolving physical states; narrative dialogue becomes continuous embodied manipulation and remote guidance; and movie-character responses become dynamic Instructor guidance. Therefore, these gains are difficult to attribute solely to movie-specific cinematography, atmosphere, or dialogue patterns. Instead, they suggest that EBM-RL learns transferable visual anchoring: identifying the current visible state, reasoning about its implications for the next interaction, and generating an appropriate response, consistent with the See-Think-Speak design.

We interpret this result cautiously. The 500-sample HoloAssist evaluation is a pilot rather than comprehensive evidence of robustness across arbitrary visual domains. Nevertheless, it provides direct supplementary evidence that EBM-RL transfers from cinematic scenes to substantially different egocentric real-world physical interactions.

\end{document}